\documentclass[sigconf]{acmart}

\AtBeginDocument{%
  \providecommand\BibTeX{{%
    \normalfont B\kern-0.5em{\scshape i\kern-0.25em b}\kern-0.8em\TeX}}}

\usepackage{times}
\usepackage{soul}
\usepackage{url}
\usepackage{caption}
\usepackage{graphicx}
\usepackage{amsmath}
\usepackage{amsthm}
\usepackage{booktabs}
\usepackage{algorithm}
\usepackage{algpseudocode}
\usepackage{caption}
\usepackage{subcaption}
\usepackage{wrapfig}
\usepackage{xcolor}
\usepackage{amsfonts}
\usepackage{tabularx}
\usepackage{enumitem}

\newtheorem{definition}{Definition}
\newtheorem{theorem}{Theorem}
\newtheorem{assumption}{Assumption}

\newcommand{\sysname}{FairSAOML}

\makeatletter
\newcommand{\multiline}[1]{%
  \begin{tabularx}{\dimexpr\linewidth-\ALG@thistlm}[t]{@{}X@{}}
    #1
  \end{tabularx}
}
\makeatother

\copyrightyear{2022}
\acmYear{2022}
\setcopyright{acmlicensed}

\acmConference[KDD '22] {Proceedings of the 28th ACM SIGKDD Conference on Knowledge Discovery and Data Mining}{August 14--18, 2022}{Washington, DC, USA}
\acmBooktitle{Proceedings of the 28th ACM SIGKDD Conference on Knowledge Discovery and Data Mining (KDD '22), August 14--18, 2022, Washington, DC, USA}
\acmPrice{15.00}
\acmDOI{10.1145/3534678.3539420}
\acmISBN{978-1-4503-9385-0/22/08}

\begin{document}

\title{Adaptive Fairness-Aware Online Meta-Learning \\ for Changing Environments}

\author{Chen Zhao}
\authornote{Both authors contributed equally to this research.}
\email{chen.zhao@utdallas.edu}
\affiliation{%
  \institution{The University of Texas at Dallas}
  \city{Richardson, Texas}
  \country{USA}
}

\author{Feng Mi}
\authornotemark[1]
\email{feng.mi@utdallas.edu}
\affiliation{%
  \institution{The University of Texas at Dallas}
  \city{Richardson, Texas}
  \country{USA}}

\author{Xintao Wu}
\email{xintaowu@uark.edu}
\affiliation{%
  \institution{University of Arkansas}
  \city{Fayetteville, Arkansas}
  \country{USA}}

\author{Kai Jiang}
\email{kai.jiang@utdallas.edu}
\affiliation{%
  \institution{The University of Texas at Dallas}
  \city{Richardson, Texas}
  \country{USA}}

\author{Latifur Khan}
\email{lkhan@utdallas.edu}
\affiliation{%
  \institution{The University of Texas at Dallas}
  \city{Richardson, Texas}
  \country{USA}}
  
\author{Feng Chen}
\email{feng.chen@utdallas.edu}
\affiliation{%
  \institution{The University of Texas at Dallas}
  \city{Richardson, Texas}
  \country{USA}}







\renewcommand{\shortauthors}{Zhao, et al.}

\begin{abstract}
    The fairness-aware online learning framework has arisen as a powerful tool for the continual lifelong learning setting. The goal for the learner is to sequentially learn new tasks where they come one after another over time and the learner ensures the statistic parity of the new coming task across different protected sub-populations (\textit{e.g.} race and gender). A major drawback of existing methods is that they make heavy use of the \textit{i.i.d} assumption for data and hence provide static regret analysis for the framework. However, low static regret cannot imply a good performance in changing environments where tasks are sampled from heterogeneous distributions. To address the fairness-aware online learning problem in changing environments, in this paper we first construct a novel regret metric FairSAR by adding long-term fairness constraints onto a strongly adapted loss regret. Furthermore, to determine a good model parameter at each round, we propose a novel adaptive fairness-aware online meta-learning algorithm, namely \sysname{}, which is able to adapt to changing environments in both bias control and model precision. The problem is formulated in the form of a bi-level convex-concave optimization with respect to the model’s primal and dual parameters that are associated with the model’s accuracy and fairness, respectively.
Theoretic analysis provides sub-linear upper bounds for both loss regret and violation of cumulative fairness constraints. 
Our experimental evaluation on different real-world datasets with settings of changing environments suggests that the proposed \sysname{} significantly outperforms alternatives based on the best prior online learning approaches.

\end{abstract}




\maketitle

\section{Introduction}
\label{sec:intro}
    In real world, data containing bias are likely collected sequentially over time and the distribution assumptions of them may vary at some critical time. 
For example, a recent news \cite{Miller-2020-NYTimes} by \textit{New York Times} reports that systematic algorithms become increasingly more discriminative to African Americans in bank loan during COVID-19 than pre-pandemic. These algorithms are built up from a sequence of data streams collected one after another over time, where at each time decision-makings are biased on the protected race population. This reveals (1) online algorithms completely ignore the importance of learning with fairness, in which fairness is defined by the equality of a predictive utility across different sub-populations, and predictions of a model are statistically independent on protected characters (\textit{e.g.} race); (2) machine learning models make heavy use of the \textit{i.i.d} assumption but this does not hold when environment changes (\textit{e.g.} before and after the pandemic). 

To control bias over time and especially ensure group fairness across different protected sub-populations, fairness-aware online algorithms capture supervised learning problems for which fairness is a concern and they are to sequentially train predictive models free from biases. 
Specifically, the goal of such algorithms is to ensure that both static loss regret, which compares the cumulative loss of the learner to that of the best fixed action in hindsight, and the violation of the sum of fair notions sub-linearly increase in the total number of rounds \cite{zhao-KDD-2021}. Although these works achieve state-of-the-art theoretic guarantees, the metric of static regret is only meaningful for stationary environments, and low static regret does not necessarily imply a good performance in changing environments since the time-invariant comparators may behave badly \cite{zhang-2020-AISTATS}.

To address the limitation of changing environments in online learning, strongly adaptive regret \cite{Daniely-2015-ICML} and dynamic regret \cite{Zinkevich-ICML-2003} attract people's attention. Dynamic regret handles changing environments from a global prospective and it compares the cumulative loss of the learner against any sequence of comparators but allows the comparators change over time. By contrast, strongly adaptive regret takes a local perspective and it gives more attentions on short time intervals. This regret can be viewed as the maximum statistic regret over all intervals \cite{Daniely-2015-ICML}. Although several works \cite{zhang-2020-AISTATS,zhang-nips-2018,Jun-2017-AISTATS} achieve sub-linear loss regret in online learning with changing environments, they completely ignore the significance of learning with fairness, which is a crucial hallmark of human intelligence.

In this paper, we introduce a novel problem, that is \textit{fairness-aware online meta-learning in changing environments}. In this problem setting, a sequence of data batches (\textit{i.e.} tasks) are collected one after another over time and the domains of these tasks may vary. The first goal of this work is to generalize the predictive learning accuracy and model fairness to the new data domain. Secondly, both loss regret and violation of cumulative fairness constraints are minimized and sublinearly increase in time.

To this end, technically we propose a novel online learning algorithm, namely \textit{fair strongly adaptive online meta-learner} (\sysname{}). In this algorithm, a set of geometric covering intervals with different lengths are carefully designed. We determine model parameters by formulating a problem composed by two main levels: online fair interval-level and meta-level learning. Problems in two levels interplay each other with two parts of parameters: primal parameters $\boldsymbol{\theta}$ regarding model accuracy and dual parameters $\boldsymbol{\lambda}$ adjusting fairness notions. More concretely, at round $t\in[T]$, a subset of intervals are selected to activate a number of experts where each expert runs an interval-specific algorithm. An expert takes a meta-solution pair $(\boldsymbol{\theta}_{t-1},\boldsymbol{\lambda}_{t-1})$ from the previous round as input and outputs an interval-level solution $(\boldsymbol{\theta}_{t,I},\boldsymbol{\lambda}_{t,I})$ at interval $I$.
Then a meta-algorithm combines the weighted actions of all experts so that a solution pair $(\boldsymbol{\theta}_{t},\boldsymbol{\lambda}_{t})$ at round $t$ can make predictions for the next round.
The main contributions are summarized:
\begin{sloppypar}
\begin{itemize}[leftmargin=*]
    \item To the best of our knowledge, for the first time a fairness-aware online meta-learning framework in changing environment is proposed. We first introduce a novel adaptive fairness-aware regret FairSAR. Then a novel algorithm \sysname{} is proposed to find a good decision at each round. Specifically, at each round the problem is formulated as a constrained bi-level convex-concave optimization with respect to a primal-dual parameter pair.
    \item Theoretically grounded analysis justifies the efficiency and effectiveness of the proposed method by demonstrating tighter bounds $O\Big((\tau\log T)^{1/2}\Big)$ for the loss regret and $O\Big((\tau T\log T)^{1/4}\Big)$ for violation of fairness constraints.
    \item We validate the performance of our approach with state-of-the-art techniques on  real-world datasets. Our results demonstrate \sysname{} can effectively adapt both accuracy and fairness in changing environments and it shows substantial improvements over the best prior works.
\end{itemize}
\end{sloppypar}

\section{Related Work}
\label{sec:relatedwork}
    \textbf{Changing environments in online learning.} Since the pioneering work \cite{Zinkevich-ICML-2003} in online learning, numerous subsequent researches \cite{Hazan-2020-LT,xie-2020-aaai} have been developed under the assumption of stationary environment with static regret. Low static regret, however, cannot imply a good performance in changing environment due to time-invariant comparators. To address this limitation, two regret metrics, dynamic regret \cite{Zinkevich-ICML-2003} and adaptive regret \cite{Hazan-2007-ARegret}, are devised to measure the learner's performance in changing environments. To bound the general dynamic regret, the path-length of comparators \cite{Zinkevich-ICML-2003,zhang-nips-2018} is introduced and further developed.
Unlike dynamic regret, adaptive regret handles changing environments from a local perspective by focusing on comparators in short intervals. 
To reduce the time complexity of adaptive regret based online algorithms, geometric covering intervals \cite{Daniely-2015-ICML,Jun-2017-AISTATS,zhang-2020-AISTATS} and data streaming techniques \cite{gyorgy-2012-efficient} are developed. 
Although existing methods achieve state-of-the-art performance, a major drawback is that they immerse in minimizing objective functions but ignore the model fairness of prediction. 

\textbf{Fairness-aware online learning} problems assume individuals arrive one at a time and the goal of such algorithms is to train predictive models free from biases. 
From the perspective of optimization, group fairness notions are normally considered as constraints added on learning objectives.
However, when the constraints are complex, the computational burden of the projection onto constraints may be too high. 
Several closely related works, including FairFML \cite{zhao-KDD-2021}, FairGLC \cite{GenOLC-2018-NeurIPS}, FairAOGD \cite{AdpOLC-2016-ICML}, aim to improve the theoretic guarantees by relaxing the output through a simpler close-form projection. 
However, these methods are not ideal for continual lifelong learning with changing task distributions, as they assume that all samples come from the same data distribution.

\textbf{Online meta-learning} addresses the issue of learning with fast adaptation, where a meta-learner learns knowledge transfer from history tasks onto new coming ones. FTML \cite{Finn-ICML-2019} can be considered as an application of MAML \cite{Finn-ICML-2017-(MAML)} in the setting of online learning. 
FairFML \cite{zhao-KDD-2021} extends FTML by controlling bias in a online working paradigm with task-specific adaptation. 
Unfortunately, none of such techniques are devised to adapt changing environments. 

In this paper, to bridge above mentioned areas, we study the problem of fairness-aware online meta-learning to deal with changing task environments. In particular,
at each round, model parameters are determined by the proposed novel algorithm \sysname{}. This algorithm refers to ideas of dynamic programming and expert tracking techniques. Inspired by fairness-aware online learning and meta-learning, a bi-level adaptation strategy is used to accommodate changing environments and learn model with accuracy and fairness.

\section{Preliminaries}
\label{sec:preliminaries}

\subsection{Notations}
An index set of a sequence of tasks is defined as $[T]=\{1,...,T\}$. Vectors are denoted by lower case bold face letters, \textit{e.g.} the primal variables $\boldsymbol{\theta}\in\Theta$ and the dual variables $\boldsymbol{\lambda}\in\mathbb{R}^m_+$ where their $i$-th entries are $\theta_i,\lambda_i$. Vectors with subscripts of task indices, such as $\boldsymbol{\theta}_t,\boldsymbol{\lambda}_t$ where $t\in[T]$, indicate model parameters for the task at round $t$. The Euclidean $\ell_2$-norm of $\boldsymbol{\theta}$ is denoted as $||\boldsymbol{\theta}||$. Given a differentiable function $\mathcal{L}(\boldsymbol{\theta,\lambda}):\Theta\times\mathbb{R}^m_+\rightarrow\mathbb{R}$, the gradient at $\boldsymbol{\theta}$ and $\boldsymbol{\lambda}$ is denoted as $\nabla_{\boldsymbol{\theta}} \mathcal{L}(\boldsymbol{\theta,\lambda})$ and $\nabla_{\boldsymbol{\lambda}} \mathcal{L}(\boldsymbol{\theta,\lambda})$, respectively. Scalars are denoted by lower case italic letters, \textit{e.g.} $\eta>0$. 
$\prod_\mathcal{B}$ is the projection operation to the set $\mathcal{B}$. 
$[\boldsymbol{u}]_+$ denotes the projection of the vector $\boldsymbol{u}$ on the nonnegative orthant in $\mathbb{R}^m_+$, namely $[\boldsymbol{u}]_+ = (\max\{0,u_1\}),...,\max\{0,u_m\})$. 
Some important notations are listed in Table \ref{tab:notation} of Appendix \ref{sec:notations}.

\subsection{Fairness-Aware Constraints}
In general, group fairness criteria used for evaluating and designing machine learning models focus on the relationships between the protected attribute and the system output \cite{Zhao-ICDM-2019,Zhao-ICKG-1-2020,wang-2021-clear}. The problem of group unfairness prevention can be seen as a constrained optimization problem. For simplicity, we consider one binary protected attribute (\textit{e.g.} gender) in this work. However, our ideas can be easily extended to many protected attributes with multiple levels.

Let $\mathcal{Z=X\times Y}$ be the data space, where $\mathcal{X} = \mathcal{E} \cup\mathcal{S}$. Here $\mathcal{E} \subset \mathbb{R}^d$ is an input space, $\mathcal{S} = \{0,1\}$ is a protected space, and $\mathcal{Y} = \{-1,1\}$ is an output space for binary classification. 
Given a task (batch) of samples
$\{\mathbf{e}_{i}, y_{i}, s_{i}\}_{i=1}^n \in(\mathcal{E\times Y\times S})$ 
where $n$ is the number of datapoints, 
a fine-grained measurement to ensure fairness in class label prediction is to design fair classifiers by controlling the notions of fairness between protected subgroups, such as demographic parity and equality of opportunity \cite{Wu-2019-WWW,Lohaus-2020-ICML}.
\begin{definition}[Notions of Fairness \cite{Wu-2019-WWW,Lohaus-2020-ICML}]
A classifier $h:\Theta\times\mathbb{R}^d\rightarrow\mathbb{R}$ is fair when its predictions are independent of the protected attribute $\mathbf{s}=\{s_i\}_{i=1}^n$.
To get rid of the indicator function and relax the exact values, a linear approximated form of the difference between protected subgroups is defined \cite{Lohaus-2020-ICML},
\begin{align}
\label{dbc definition}
    g(\boldsymbol{\theta})=\Big|\mathbb{E}_{(\mathbf{e},y,s)\in\mathcal{Z}}\Big[\frac{1}{\hat{p}_1(1-\hat{p}_1)}\Big(\frac{s+1}{2}-\hat{p}_1\Big)h(\boldsymbol{\theta},\mathbf{e})\Big] \Big|-\epsilon
\end{align}
where $|\cdot|$ is the absolute function and $\epsilon>0$ is the fairness relaxation determined by empirical analysis. $\hat{p}_1$ is an empirical estimate of $pr_1$. $pr_1$ is the proportion of samples in group $s=1$ and correspondingly $1-pr_1$ is the proportion of samples in group $s=0$. 
\end{definition}
Notice that, in Definition \ref{dbc definition}, when $\hat{p}_1=\mathbb{P}_{(\mathbf{e},y,s)\in\mathcal{Z}}(s=1)$, the fairness notion $g(\boldsymbol{\theta})$ is defined as the difference of demographic parity (DDP). Similarly, when $\hat{p}_1=\mathbb{P}_{(\mathbf{e},y,s)\in\mathcal{Z}}(y=1, s=1)$, $g(\boldsymbol{\theta})$ is defined as the difference of equality of opportunity (DEO) \cite{Lohaus-2020-ICML}.
Therefore, parameters $\boldsymbol{\theta}$ in the domain of a task is feasible if it satisfies the fairness constraint $g(\boldsymbol{\theta})\leq 0$. 

\subsection{Fairness-Aware Online Learning}
\label{sec:fairness-aware online learning}
The protocol of fairness-aware online convex optimization can be viewed as a repeated game between a learner and an adversary, where the learner is faced with tasks $\{\mathcal{D}_t\}_{t=1}^T$ one after another. At each round $t\in[T]$, 
\begin{itemize}[leftmargin=*]
    \item \textbf{Step 1}: The learner selects a model parameter $\boldsymbol{\theta}_t$ in the fair domain $\Theta$.
    \item \textbf{Step 2}: The adversary reveals a loss function $f_t:\Theta\times\mathbb{R}^d\rightarrow\mathbb{R}$ and $m$ fairness functions $g_i:\Theta\times\mathbb{R}^d\rightarrow\mathbb{R},\forall i\in[m]$.
    \item \textbf{Step 3}: The learner incurs an instantaneous loss $f_t(\boldsymbol{\theta}_t,\mathcal{D}_t)$ and $m$ fairness notions $g_i(\boldsymbol{\theta}_t,\mathcal{D}_t),\forall i\in[m]$. 
    \item \textbf{Step 4}: Advance to round $t+1$.
\end{itemize}


The goal of fairness-aware online learning \cite{GenOLC-2018-NeurIPS,zhao-KDD-2021} is to (1) minimize the loss regret over the rounds which is to compare to the cumulative loss of the best fixed model in hindsight, and (2) ensure the total violation of fair constraints sublinearly increase in $T$. The loss regret is typically referred as \textit{static regret} since the comparator is time-invariant. To control bias and especially ensure group fairness across different protected sub-populations, fairness notions are considered as constraints added on optimization problems.
\begin{align}
\label{eq:static-regret}
    \min_{\boldsymbol{\theta}_1,...,\boldsymbol{\theta}_T\in\Theta} \quad & \text{Regret}(T) = \sum_{t=1}^T f_t(\boldsymbol{\theta}_t,\mathcal{D}_t)
    - \min_{\boldsymbol{\theta}\in\Theta}\sum_{t=1}^T f_t(\boldsymbol{\theta},\mathcal{D}_t) \\
    \text{subject to} \quad &\sum_{t=1}^T g_i(\boldsymbol{\theta}_t,\mathcal{D}_t) \leq O(T^\gamma), \forall i\in[m], \gamma\in(0,1) \nonumber
\end{align}
where the summation of fair constraints is defined as \textit{long-term constraints} in \cite{OGDLC-2012-JMLR}. 
The big $O$ notation in the constraint is to bound the total violation of fairness sublinear in $T$.
The main drawback of using the metric of static regret is that it is only meaningful
for stationary environments, and low static regret cannot imply a good performance in changing environments since the time-invariant comparator in Eq.(\ref{eq:static-regret}) may behave badly \cite{zhang-2020-AISTATS}.

\begin{figure*}[t!]
  \begin{center}
    \includegraphics[width=0.8\textwidth]{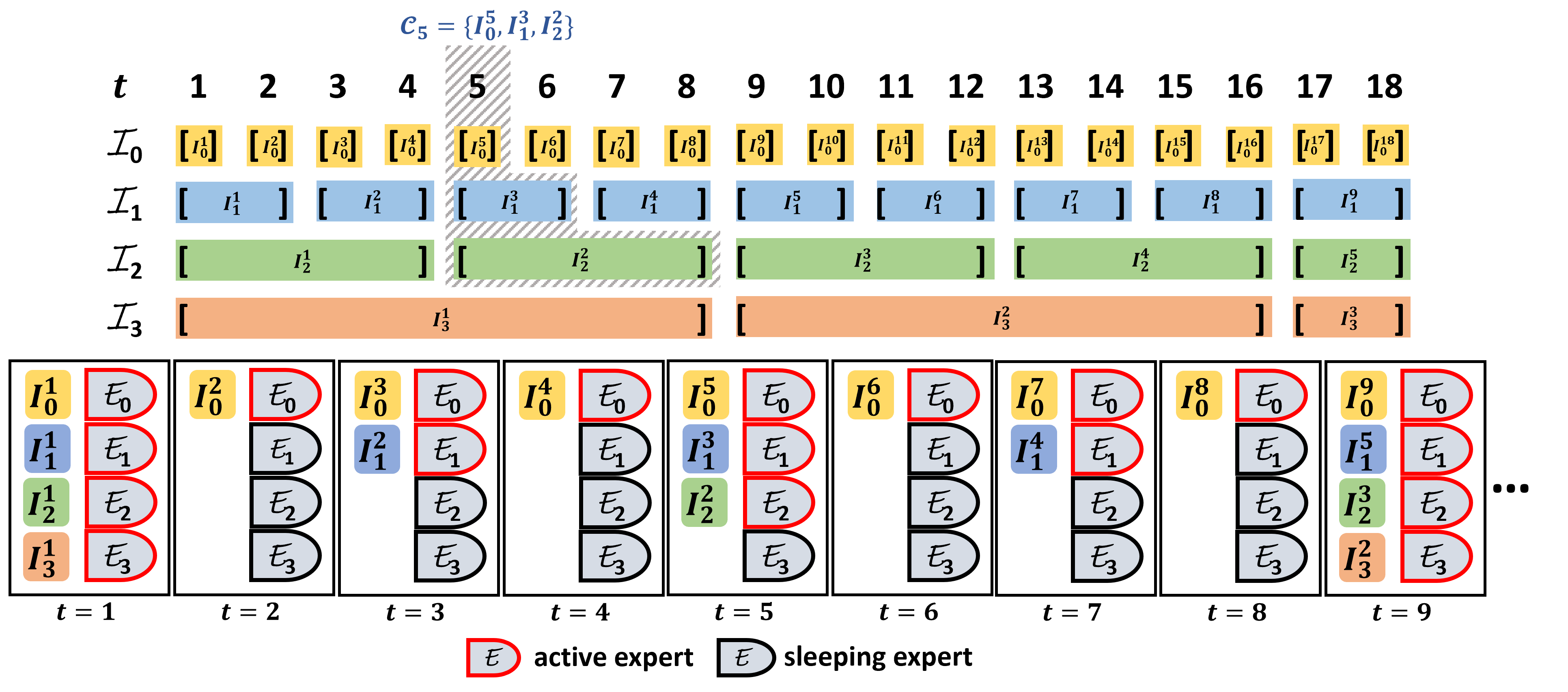}
    \vspace{-5mm}
  \end{center}
  \caption{(Upper) An graphical illustration of AGC intervals (base=2) when $T=18$. The total interval set $\mathcal{I}$ consists of 4 subsets $\{\mathcal{I}_0,\mathcal{I}_1,\mathcal{I}_2,\mathcal{I}_3\}$ and each contains different numbers of intervals with fixed length. Intervals covered by shadow is an example of target subset $\mathcal{C}_5$ when $t=5$. (Lower) A demonstration of active experts activated by $\mathcal{C}_t$ and their corresponding sleeping experts.}
\label{fig:AGCandExamples}
\vspace{-5mm}
\end{figure*}

\section{Methodology}
\label{sec:methodology}
    \subsection{Settings and Problem Formulation}
\label{sec:settings and problem formulation}
To address the limitation of changing environments in online learning, \textit{adaptive regret} (AR) based on \cite{Hazan-2007-ARegret} is defined as the maximum static regret over any contiguous intervals. However, AR does not respect short intervals well. To this end, \textit{strongly adaptive regret} (SAR) \cite{daniely15-2015-ICML} is proposed to improve AR, which emphasizes the dependence on lengths of intervals and it takes the form that
\begingroup\makeatletter\def\f@size{7.95}\check@mathfonts
\def\maketag@@@#1{\hbox{\m@th\large\normalfont#1}}%
\begin{align}
\label{eq:SAR}
    \text{SAR}(T,\tau) = \max_{[s,s+\tau-1]\subseteq[T]} \Big ( \sum_{t=s}^{s+\tau-1} f_t(\boldsymbol{\theta}_t,\mathcal{D}_t) - \min_{\boldsymbol{\theta}\in\Theta}\sum_{t=s}^{s+\tau-1} f_t(\boldsymbol{\theta},\mathcal{D}_t) \Big )
\end{align}
\endgroup
where $\tau$ indicates the length of time interval. In SAR, the learner is competing with changing comparators, as $\boldsymbol{\theta}$ varies with $s$ over $[s, s+\tau-1]$. 

In this paper, we consider the online meta-learning setting that is similar in \cite{Finn-ICML-2019,zhao-KDD-2021} but tasks are sampled from heterogeneous distributions. Instead of static regret, we define a novel regret FairSAR in Eq.(\ref{eq:ourRegret}). Let $\{\boldsymbol{\theta}_t\}_{t=1}^T$ be the sequence of model parameters generated in the \textbf{Step 1} of the learning protocol (see Sec. \ref{sec:fairness-aware online learning}).
The goal of our problem is to minimize FairSAR under the long-term fair constraints:
\begingroup\makeatletter\def\f@size{8}\check@mathfonts
\def\maketag@@@#1{\hbox{\m@th\large\normalfont#1}}%
\begin{align}
\label{eq:ourRegret}
    &\text{FairSAR}(T,\tau)=\max_{[s,s+\tau-1]\subseteq[T]} \bigg( \sum_{t=s}^{s+\tau-1} f_t\Big(\mathcal{G}_t(\boldsymbol{\theta}_t,\mathcal{D}_t^S),\mathcal{D}_t^V\Big)\nonumber\\
    & \quad \quad \quad \quad \quad \quad \quad \quad \quad \quad \quad - \min_{\boldsymbol{\theta}\in\Theta}\sum_{t=s}^{s+\tau-1} f_t\Big(\mathcal{G}_t(\boldsymbol{\theta},\mathcal{D}_t^S),\mathcal{D}_t^V\Big) \bigg) \\
    &\text{subject to} \max_{[s,s+\tau-1]\subseteq[T]} \bigg (\sum_{t=s}^{s+\tau-1} g_i\Big(\mathcal{G}_t(\boldsymbol{\theta}_t,\mathcal{D}_t^S),\mathcal{D}_t^V\Big) \bigg)\leq O(T^\gamma), \forall i \in[m] \nonumber
\end{align}
\endgroup
where $\gamma\in(0,1)$. 
$\mathcal{D}_t^S, \mathcal{D}^{V}_t\subset\mathcal{D}_t$ are the support and validation set.
$\mathcal{G}_t(\cdot)$ is the base learner which corresponds to one or multiple gradient steps \cite{Finn-ICML-2017-(MAML)}.
Different from traditional online learning settings, the long-term constraint violation $g(\cdot): \mathcal{B}\times\mathbb{R}^d\rightarrow\mathbb{R}$ is satisfied. To facilitate our analysis, $\boldsymbol{\theta}_t$ is originally chosen from its domain
$\Theta=\{\boldsymbol{\theta}\in\mathbb{R}^d:g_i(\boldsymbol{\theta},\mathcal{D}_t)\leq 0, \forall i\in[m]\}$. A projection operator is hence typically applied to the updated variables in order to make them feasible \cite{OGDLC-2012-JMLR,GenOLC-2018-NeurIPS,AdpOLC-2016-ICML}.
In order to lower the computational complexity and accelerate the online processing speed, we relax the domain $\Theta$ to $\mathcal{B}$, where $\Theta\subseteq\mathcal{B}=S\mathbb{K}$ with $\mathbb{K}$ being the unit $\ell_2$ ball centered at the origin, and $S=\max\{r>0: r=||\boldsymbol{\theta}_1-\boldsymbol{\theta}_2||, \forall \boldsymbol{\theta}_1,\boldsymbol{\theta}_2\in\Theta \}$. 

In the protocol stated in Sec. \ref{sec:fairness-aware online learning}, the key step (\textbf{Step 1}) is to find a good parameter $\boldsymbol{\theta}_t$ at each round.
Let $\mathcal{K}$ be an online learning algorithm determining $\boldsymbol{\theta}_t$. A trick commonly used in designing a meta algorithm for changing environments is to initiate a new instance of $\mathcal{K}$ at every round \cite{Hazan-2007-ARegret}. 
Therefore, we run $\mathcal{K}$ independently for each interval $I\in\{[t...\infty]|t=1,2,...\}$ and  denote $\mathcal{K}_I$ as the run of $\mathcal{K}$ on interval $I$. At round $t$, we combine the decisions from the runs by weighted average. 
The key idea is that at time $t$, some of the outputs in $\mathcal{K}\in\{\mathcal{K}_I\}_{I\ni t}$ are not based on any data prior to time $t'<t$, so that if the environment changes at $t'$, those outputs may be given a larger weight by the meta algorithm, allowing it to adapt more quickly to the change. 
A main drawback, however, is a factor of $t$ increase in the time complexity. To avoid this, we reduce it to $O(\log t)$ by restarting algorithms on a designed set of adaptive geometric covering (AGC) intervals.

\subsection{AGC Intervals and Experts}
\label{sec:agc intervals and experts}
In Eq.(\ref{eq:ourRegret}), FairSAR evaluates the learner's performance on each time interval and it is the maximum regret over any contiguous intervals. 
Similar to the seminal work of SAR \cite{Daniely-2015-ICML}, we construct a number of interval sets, namely adaptive geometric covering (AGC) intervals, where each set contains various intervals with same length. 
AGC intervals can be considered as a special case of a more general set of intervals and they hence efficiently reduce time complexity to $O(\log t)$. A set of contiguous AGC intervals $\mathcal{I}$ are defined as
\begin{align}
\label{eq:AGC}
    \mathcal{I} = \bigcup_{k\in[\lfloor\log_2^T\rfloor-1]\cup \{0\}} \mathcal{I}_k
\end{align}
where $\forall k, \mathcal{I}_k=\{I_k^i|[(i-1)\cdot 2^k+1, \min{\{T,i\cdot 2^k\}}]:i\in\mathbb{N}\}$. An example with $T=18$ is given in Figure \ref{fig:AGCandExamples} to illustrate the composition of AGC intervals. Notice that the $\log$ base equals to $2$ is not required, but larger base number leads to less interval levels. 

Furthermore, inspired by \textit{learning with expert advice} problems \cite{Jun-2017-AISTATS}, we construct a set $\mathcal{U}$ of interval-level learning processes, defined as experts, where $|\mathcal{U}|=\lfloor\log_2^T\rfloor$. 
To better adapt changing environment, only one expert is assigned to the corresponding interval subset $\mathcal{I}_k$. At each round $t\in[T]$, we introduce a target set $\mathcal{C}_t\subset \mathcal{I}$ which includes a set of intervals starting from $t$
\begin{align}
\label{eq:C_t}
    \mathcal{C}_t = \{I|I\in\mathcal{I},t\in I,(t-1)\notin I\}
\end{align}
$\mathcal{C}_t$ aims to dynamically activate a subset of experts in $\mathcal{U}$ (an example of $C_{t=5}$ is given in Figure \ref{fig:AGCandExamples}). A set of active experts at round $t$ is denoted as $\mathcal{A}_t\subseteq\mathcal{U}$ and the corresponding inactive ones are called sleeping experts. The set of sleeping experts is denoted $\mathcal{S}_t=\mathcal{U}\setminus\mathcal{A}_t$. Note that only active experts are used to learn and further update interval-level model parameters through a base learner $\mathcal{G}_t(\cdot)$ that corresponds to one or multiple gradient steps \cite{Finn-ICML-2017-(MAML)}. The \textit{interval-level} parameter update for an active expert $E_I$ in $\mathcal{A}_t$ at round $t$ is defined
\begin{align}
\label{eq:inner-problem}
    \boldsymbol{\theta}_{t,I} := \mathcal{G}_{t}(\boldsymbol{\theta},\mathcal{D}^{S}_{t,I}) = &\arg\min_{\boldsymbol{\theta}} \quad f_{t}(\boldsymbol{\theta},\mathcal{D}^{S}_{t,I}) \\
    &\text{subject to} \quad g_{i}(\boldsymbol{\theta},\mathcal{D}^{S}_{t,I})\leq 0, \forall i \in [m] \nonumber 
\end{align}
where the loss function $f_{t}(\cdot)$ and the fairness function $g_{i}(\cdot)$ are defined based on the support set $\mathcal{D}_{t,I}^S\subset\mathcal{D}_{t,I}$ associated with $E_I$.




\begin{figure*}
    \centering
    \includegraphics[width=0.8\linewidth]{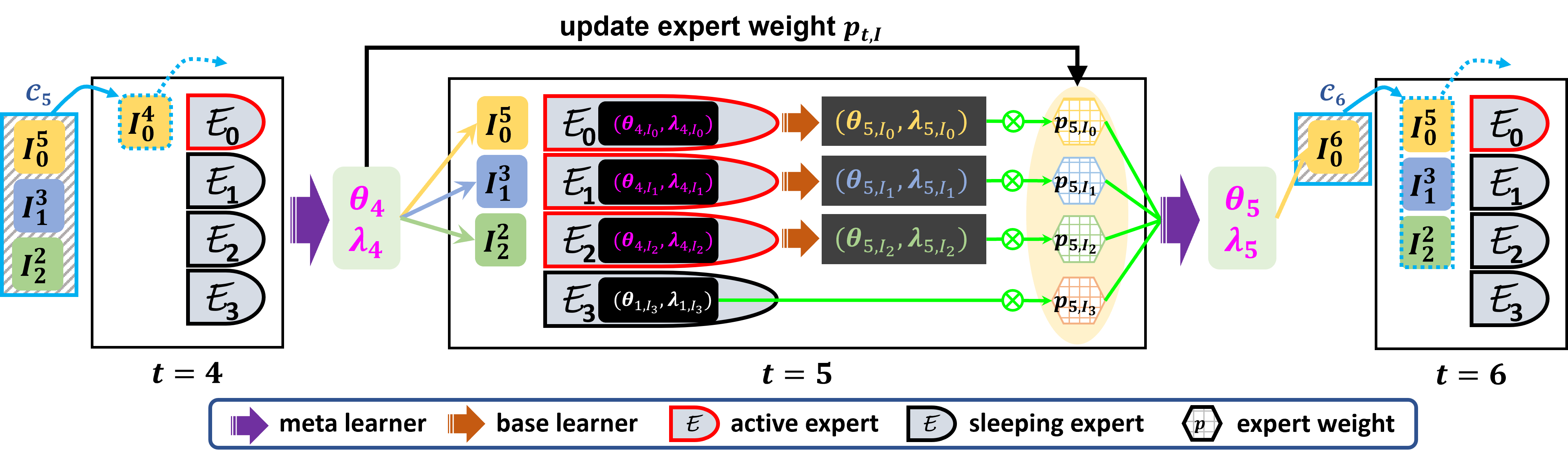}
    \vspace{-3mm}
    \caption{An overview of our \sysname{} to determine model parameter pair at each round. A target set (shadowed) of intervals are initially selected and are later used to activate corresponding experts. Each active expert runs through a base learner for the interval-level parameter-pair adaption and its weight is updated. The meta-level parameter-pair is finally attained through the meta learner by combining weighted actions of all experts.}
    \label{fig:overview}
    \vspace{-5mm}
\end{figure*}

\subsection{Learning Dynamically for Bi-Level Adaptation}
Recall that in the protocol of fairness-aware online learning (Sec. \ref{sec:fairness-aware online learning}), the main goal for the learner is to sequentially decide on model parameter $\boldsymbol{\theta}_t$ that performs well on the loss sequence and the long-term fair constraints. Crucially, inspired by \cite{Finn-ICML-2017-(MAML)}, we consider a setting where at each round $t$ the learner can perform a number of expert-specific updates at an interval level in the active set $\mathcal{A}_t$.

As specified in Eq.(\ref{eq:ourRegret}), model parameters at each round $t$ are determined by formulating problems with a nested bi-level adaptation process: interval-level and meta-level. Each level corresponds to a sub-learner, \textit{i.e.} base and meta learner respectively, described in Figure \ref{fig:overview}.
The problem of learning a meta-level parameter $\boldsymbol{\theta}_t$ is embedded with the optimization problem of finding interval-level parameters $\boldsymbol{\theta}_{t,I}$ in Eq.(\ref{eq:inner-problem}). For experts in the sleeping set $\mathcal{S}_t$, the base learner is not applied.
The \textit{meta-level} problem takes the form in Eq.(\ref{eq:outer-problem}). For simplicity, we abuse the subscription $E_I$ with $I$.
\begingroup\makeatletter\def\f@size{9}\check@mathfonts
\def\maketag@@@#1{\hbox{\m@th\large\normalfont#1}}%
\begin{align}
\label{eq:outer-problem}
    &\min_{\boldsymbol{\theta}\in\mathcal{B}} \sum_{E_I\in\mathcal{A}_t} p_{t,I} \cdot f_{t}(\mathcal{G}_{t}(\boldsymbol{\theta},\mathcal{D}_{t,I}^S),\mathcal{D}_{t,I}^Q) + \sum_{E_I\in\mathcal{S}_t} p_{t,I}\cdot f_{t}(\boldsymbol{\theta}_{t',I},\mathcal{D}_{t,I}^Q) \\
    &\text{s.t.} \sum_{E_I\in\mathcal{A}_t} p_{t,I} \cdot g_i(\mathcal{G}_{t}(\boldsymbol{\theta},\mathcal{D}_{t,I}^S),\mathcal{D}_{t,I}^Q) + \sum_{E_I\in\mathcal{S}_t} p_{t,I}\cdot g_i(\boldsymbol{\theta}_{t',I},\mathcal{D}_{t,I}^Q)\leq 0 \nonumber
\end{align}
\endgroup
where $p_{t,I}\geq 0$ is the expert weight of $E_I$ at $t$. $\mathcal{D}_{t,I}^Q\subset\mathcal{D}_{t,I}$ is the query set where $\mathcal{D}_{t,I}^Q\cap\mathcal{D}_{t,I}^S=\emptyset$.
$\boldsymbol{\theta}_{t',I}$ is the interval-level model parameter for an sleeping expert $E_I\in\mathcal{S}_t$ where the round index $t'<t$ represents the last time this expert was activated.

In the following section, we introduce our proposed algorithm \sysname{}. In stead of optimizing primal parameters only, it efficiently deals with the bi-level optimization problem of Eq.(\ref{eq:inner-problem})(\ref{eq:outer-problem}) by approximating a sequence of pairs of primal-dual meta parameters $\{(\boldsymbol{\theta}_t,\boldsymbol{\lambda}_t)\}_{t=1}^T$ where the pair respectively responds for adjusting accuracy and fairness level.

\subsection{An Efficient Algorithm: \sysname{}}
To find a good model parameter pair $(\boldsymbol{\theta}_t, \boldsymbol{\lambda}_t)$ at each round, an efficient working flow is proposed in Algorithm \ref{alg:ouralgorithm} and an overview of our proposed \sysname{} is given in Figure \ref{fig:overview}. Inspired by dynamic programming and expert-tracking \cite{luo-2015-achieving} techniques, experts at each round are recursively divided into active and sleeping sets. Model parameters in active experts are locally updated, but for the ones in sleeping experts, they are directly inherited from the previous round.
Specifically, in the beginning of round $t$, a target set $C_t$ containing intervals is used to activate a subset of experts in $\mathcal{U}$. For each active expert $E_I$ in $\mathcal{A}_t$, an interval-level algorithm takes the meta-level solution $(\boldsymbol{\theta}_{t-1}, \boldsymbol{\lambda}_{t-1})$ and outputs an expert-specific solution pair $(\boldsymbol{\theta}_{t,I}, \boldsymbol{\lambda}_{t,I})$. 
Finally, through the meta-learner, we combine the weighted solutions of all experts and move to the next round.

We explain the main steps in Algorithm \ref{alg:ouralgorithm} below. In step 4, when a new task arrives at round $t$, a batch of data $\mathcal{D}_t^{V}$ is randomly sampled from $\mathcal{D}_t$ for validation purpose and the performance on $\boldsymbol{\theta}_{t-1}$ achieved is recorded. A target set of intervals $\mathcal{C}_t$ is selected from $\mathcal{I}$ in step 5. 
For each interval $I \in\mathcal{C}_t$ (step 6-11), the corresponding expert $E_{t,I}$ is activated.
We set adaptive stepsizes $\eta_{t,I}=S/(G\sqrt{|I|})$ where $|I|=2^k$ denotes the interval length in $\mathcal{I}_k$. $S$ is the radius of the Euclidean ball $\mathcal{B}$ and there exists a constant $G>0$ that bounds the (sub)gradients of $f_t$ and $g_i$. Following the setting used in \cite{AdpOLC-2016-ICML}, empirically we set $S=\sqrt{1+2\epsilon}-1$ and $G=\max\{\sqrt{d}+S,\max_t\{||\boldsymbol{e}_I||_2, \boldsymbol{e}_I\in\mathcal{P}_t\} \}$, where $\boldsymbol{e}_I$ is the non-protected features lied in the interval $I$ and $d$ is its feature dimension. $\mathcal{P}_t$ is a set which includes all past intervals until time $t$. 
In steps 12-14, for all experts in $\mathcal{U}$, a following weight $p_{t,I}$ is assigned:
\begin{align}
\label{weight}
    p_{t,I} = \frac{w(R_{t,I}, C_{t,I})}{\sum_{E_I\in\mathcal{U}}w(R_{t,I}, C_{t,I})}
\end{align}
Here, a weight function \cite{luo-2015-achieving} is defined 
as $w(R,C) = \frac{1}{2}\big(\Phi(R+1,C+1)-\Phi(R-1,C-1)\big)$, where $\Phi(R,C) = \exp([R]^2_+/3C)$
and $[r]_+=\max(0,r)$ and $\Phi(0,0)=1$.
In steps 15-24, our \sysname{} responds to the bi-level adaptation stated in Eq.(\ref{eq:inner-problem}) and (\ref{eq:outer-problem}). Specifically, to solve the interval-level problem in Eq.(\ref{eq:inner-problem}), for each active expert $E_I$ in $\mathcal{A}_t$, we consider following Lagrangian function
\begingroup\makeatletter\def\f@size{8.5}\check@mathfonts
\def\maketag@@@#1{\hbox{\m@th\large\normalfont#1}}%
\begin{align}
    \mathcal{F}_{t,I}(\boldsymbol{\theta}_{t-1},\boldsymbol{\lambda}_{t-1})
    =f_{t}(\boldsymbol{\theta}_{t-1},\mathcal{D}_{t,I}^S) + \sum_{i=1}^m \lambda_{t-1,i}\cdot g_{i}(\boldsymbol{\theta}_{t-1},\mathcal{D}_{t,I}^S)
\end{align}
\endgroup
where the interval-level parameter pair for an active expert $E_I$ are initialized with the meta-level parameter $(\boldsymbol{\theta}_{t-1}, \boldsymbol{\lambda}_{t-1})$ . 
For optimization with simplicity, cumulative constraints in Eq.(\ref{eq:inner-problem}) are approximated with the summarized regularization. Interval-level parameters are updated through a base learner $\mathcal{G}_t(\cdot)$. One example for the learner is updating with one gradient step \cite{Finn-ICML-2017-(MAML)} using the pre-determined adaptive stepsize $\eta_{t,I}$. Notice that for multiple gradient steps, $\boldsymbol{\theta}_{t,I}$ and $\boldsymbol{\lambda}_{t,I}$ interplay each other for updating.
\begin{align}
\label{eq:inner-pd-update}
    &\boldsymbol{\theta}_{t,I} = \boldsymbol{\theta}_{t-1} - \eta_{t,I}\nabla_{\boldsymbol{\theta}}\mathcal{F}_{t,I}(\boldsymbol{\theta}_{t-1},\boldsymbol{\lambda}_{t-1}) \\
    &\boldsymbol{\lambda}_{t,I} = \boldsymbol{\lambda}_{t-1} + \eta_{t,I}\nabla_{\boldsymbol{\lambda}}\mathcal{F}_{t,I}(\boldsymbol{\theta}_{t,I},\boldsymbol{\lambda}_{t-1})
\end{align}
Next, to solve the meta-level problem in Eq.(\ref{eq:outer-problem}), we combine the actions of active experts together with sleeping experts. 
We consider the following augmented Lagrangian function and abuse the symbol $t'$ with $t$ in Eq.(\ref{eq:outer-problem}):
\begingroup\makeatletter\def\f@size{8.5}\check@mathfonts
\def\maketag@@@#1{\hbox{\m@th\large\normalfont#1}}%
\begin{align}
\label{eq:L_t}
    &\mathcal{L}_t(\boldsymbol{\theta}_{t,I},\boldsymbol{\lambda}_{t,I}) = \sum_{E_I\in\mathcal{U}} p_{t,I} \Bigg( f_{t}(\boldsymbol{\theta}_{t,I},\mathcal{D}_{t,I}^Q) \\
    &\quad\quad\quad\quad\quad\quad\quad\quad + \sum_{i=1}^m \Big(\lambda_{i,t,I}\cdot g_i(\boldsymbol{\theta}_{t,I},\mathcal{D}_{t,I}^Q) 
    - \frac{\delta(\eta_1+\eta_2)}{2}\lambda_{i,t,I}^2 \Big) \Bigg) \nonumber
\end{align}
\endgroup
where $\delta > 0$ is a constant determined by analysis. 
Note that the last augmented term on the dual variable is devised to prevent $\boldsymbol{\lambda}$ from being too large. The update rule for meta-level parameters follows:
\begin{align}
\label{eq:outer-pd-update}
    \boldsymbol{\theta}_{t} 
    &= \prod_\mathcal{B}\Big(\boldsymbol{\theta}_{t-1} - \eta_1\nabla_{\boldsymbol{\theta}}\mathcal{L}_t(\boldsymbol{\theta}_{t,I},\boldsymbol{\lambda}_{t,I})\Big) \\
    \boldsymbol{\lambda}_{t} &= \Big[ \boldsymbol{\lambda}_{t-1} + \eta_2\nabla_{\boldsymbol{\lambda}}\mathcal{L}_t(\boldsymbol{\theta}_{t,I},\boldsymbol{\lambda}_{t,I}) \Big]_+
\end{align}
where $\prod_\mathcal{B}$ is the projection operation to the relaxed domain $\mathcal{B}$ that is introduced in Sec.\ref{sec:settings and problem formulation}. This approximates the true desired projection with a simpler closed-form.
Finally, in steps 25-28, for each expert we update its $R$ and $C$ values which determine the expert weight for the next round. The intuition of weight update is to re-adjust difference between the meta-solution and the interval-level solution given by the expert.

\begin{algorithm}[!t]
\caption{\sysname{}}
\label{alg:ouralgorithm}
\begin{algorithmic}[1]
\State Initialize meta-parameters pair $(\boldsymbol{\theta}_0, \boldsymbol{\lambda}_0)$, where $\boldsymbol{\theta}_0$ is the center of $\mathcal{B}$ and $\boldsymbol{\lambda}_0\in\mathbb{R}_+^m$ is randomly chosen
\State \multiline{%
    Initialize $k=\lfloor\log_2^T\rfloor$ experts in set $\mathcal{U}$ and for each expert $E_I\in\mathcal{U}$, set $R_{0,I}=0, C_{0,I}=0, p_{0,I}=1$.}
\For{each $t\in[T]$}
    \State \multiline{%
        Sample $\mathcal{D}^{V}_t\subset\mathcal{D}_t$ and record the performance of $\boldsymbol{\theta}_{t-1}$}
    \State Subset $\mathcal{C}_t$ from $\mathcal{I}$ using Eq.(\ref{eq:C_t})
    \For{each interval $I\in\mathcal{C}_t$}
        \State \multiline{%
            Activate expert $E_{t,I}$ and set $\eta_{t,I}=S/(G\sqrt{|I|})$, $(\boldsymbol{\theta}_{t,I},\boldsymbol{\lambda}_{t,I})\leftarrow(\boldsymbol{\theta}_{t-1},\boldsymbol{\lambda}_{t-1})$}
        \State Identify an $E_J\in\mathcal{U}$, such that $|J|=|I|$
        \State Update $R_{t,I}\leftarrow R_{t-1,J}$ and $C_{t,I}\leftarrow C_{t-1,J}$
        \State Replace $E_J$ with $E_I$ in $\mathcal{U}$
    \EndFor
    \For{each expert in $\mathcal{U}$}
        \State Update $p_{t,I}$ using $R_{t,I}, C_{t,I}$ in Eq.(\ref{weight})
    \EndFor
    \For{$n=1,...,N_{meta}$ steps}
        \For{each active expert $E_I$ in $\mathcal{A}_t$}    
            \State Sample support set $\mathcal{D}_{t,I}^{S}\subset\mathcal{D}_{t,I}$
            \State \multiline{%
                Adapt interval-level primal and dual variables with $\mathcal{D}_{t,I}^{S}$ using Eq.(\ref{eq:inner-pd-update})(12)}
        \EndFor
        \For{each expert $E_I$ in $\mathcal{U}$}
            \State Sample query set $\mathcal{D}_{t,I}^{Q}\subset\mathcal{D}_{t,I}$
        \EndFor
        \State \multiline{%
            Update meta-level primal and dual variables with $D^Q_{t,I}$ using Eq.(\ref{eq:outer-pd-update})(15)}
    \EndFor
    \For{each expert $E_I\in\mathcal{U}$}
        \State $R_{t+1,I} = R_{t,I} + \mathcal{F}_{t,I}(\boldsymbol{\theta}_t,\boldsymbol{\lambda}_t) - \mathcal{F}_{t,I}(\boldsymbol{\theta}_{t,I},\boldsymbol{\lambda}_{t,I})$
        \State $C_{t+1,I} = C_{t,I} + \Big | \mathcal{F}_{t,I}(\boldsymbol{\theta}_t,\boldsymbol{\lambda}_t) - \mathcal{F}_{t,I}(\boldsymbol{\theta}_{t,I},\boldsymbol{\lambda}_{t,I})\Big |$
    \EndFor
\EndFor
\end{algorithmic}
\end{algorithm}
\setlength{\textfloatsep}{3pt}

\section{Analysis}
\label{sec:analysis}
    To analysis, we first make following assumptions as in \cite{zhang-2020-AISTATS,OGDLC-2012-JMLR}. 
Examples where these assumptions hold include logistic regression and $L2$ regression over a bounded domain. As for constraints, a family of fairness notions, such as DDP stated in Definition \ref{dbc definition}, are applicable as discussed in \cite{Lohaus-2020-ICML}. For simplicity, in this section we omit $\mathcal{D}$ used in $f_t(\cdot),\forall t$ and $g_i(\cdot),\forall i$.

\begin{assumption}[Convex domain]
\label{assmp1}
    The convex set $\Theta$ is non-empty, closed, bounded, and it is described by $m$ convex functions as $\Theta = \{\boldsymbol{\theta}:g_i(\boldsymbol{\theta})\leq 0, \forall i\in[m]\}$. The relaxed domain $\mathcal{B}$ (where $\Theta\subseteq\mathcal{B}$) contains the origin $\boldsymbol{0}$ and its diameter is bounded by $S$.
\end{assumption}

\begin{assumption}
Both the loss functions $f_t(\cdot), \forall t$ and constraint functions $g_i(\cdot), \forall i\in[m]$ satisfy the following assumptions 
\label{assmp2}
\begin{enumerate}[leftmargin=*]
    \item
    \begin{sloppypar}
    (Lipschitz Continuous) $\forall \boldsymbol{\theta}_1,\boldsymbol{\theta}_2\in\mathcal{B}$,  $||f_t(\boldsymbol{\theta}_1)-f_t(\boldsymbol{\theta}_2)||\leq L_f||\boldsymbol{\theta}_1-\boldsymbol{\theta}_2||, ||g_i(\boldsymbol{\theta}_1)-g_i(\boldsymbol{\theta}_2)||\leq L_g||\boldsymbol{\theta}_1-\boldsymbol{\theta}_2||$. 
    Let $G = \max\{L_f,L_g\}$, $F = \max_{t\in[T]}\max_{\boldsymbol{\theta}_1,\boldsymbol{\theta}_2\in\mathcal{B}} f_t(\boldsymbol{\theta}_1)-f_t(\boldsymbol{\theta}_2)\leq 2L_f S$, and $D = \max_{i\in[m]}\max_{\boldsymbol{\theta}\in\mathcal{B}}g_i(\boldsymbol{\theta})\leq L_g S$
    
    \item (Lipschitz Gradient) $f_t(\boldsymbol{\theta}), \forall t$ are $\beta_f$-smooth and $g_i(\boldsymbol{\theta}),\forall i$ are $\beta_g$-smooth, that is, $\forall \boldsymbol{\theta}_1,\boldsymbol{\theta}_2\in\mathcal{B}$, $||\nabla f_t(\boldsymbol{\theta}_1)-\nabla f_t(\boldsymbol{\theta}_2)||\leq \beta_f||\boldsymbol{\theta}_1-\boldsymbol{\theta}_2||, ||\nabla g_i(\boldsymbol{\theta}_1)-\nabla g_i(\boldsymbol{\theta}_2)||\leq \beta_g||\boldsymbol{\theta}_1-\boldsymbol{\theta}_2||$.
    
    \item (Lipschitz Hessian) Twice-differentiable functions $f_t(\boldsymbol{\theta}), \forall t$ and $g_i(\boldsymbol{\theta}),\forall i$ have $\rho_f$ and $\rho_g$- Lipschitz Hessian, respectively. That is, $\forall \boldsymbol{\theta}_1-\boldsymbol{\theta}_2\in\mathcal{B}$, $||\nabla^2 f_t(\boldsymbol{\theta}_1)-\nabla^2 f_t(\boldsymbol{\theta}_2)||\leq \rho_f||\boldsymbol{\theta-\phi}||, ||\nabla^2 g_i(\boldsymbol{\theta}_1)-\nabla^2 g_i(\boldsymbol{\theta}_2)||\leq \rho_g||\boldsymbol{\theta}_1-\boldsymbol{\theta}_2||$.
    \end{sloppypar}
\end{enumerate}
\end{assumption}

\begin{assumption}[Strongly convexity]
\label{assmp3}
    \begin{sloppypar}
    Suppose $f_t(\boldsymbol{\theta}), \forall t$ and $g_i(\boldsymbol{\theta}),\forall i$ have strong convexity, that is, $\forall \boldsymbol{\theta}_1, \boldsymbol{\theta}_2\in\mathcal{B}$, $||\nabla f_t(\boldsymbol{\theta}_1)-\nabla f_t(\boldsymbol{\theta}_2)||\geq \mu_f||\boldsymbol{\theta}_1-\boldsymbol{\theta}_2||, ||\nabla g_i(\boldsymbol{\theta}_1)-\nabla g_i(\boldsymbol{\theta}_2)||\geq \mu_g||\boldsymbol{\theta}_1-\boldsymbol{\theta}_2||$.
    \end{sloppypar}
\end{assumption}

Under above assumptions, we state the key Theorem \ref{main theorem} that the proposed \sysname{} enjoys sub-linear guarantee for both regret and long-term fairness constraints in the long run for Algorithm \ref{alg:ouralgorithm}. 
Proof is given in Appendix \ref{App:proof}.

\begin{theorem}
\label{main theorem}
\begin{sloppypar}
    Set $\boldsymbol{\theta}^*=\arg\min_{\boldsymbol{\theta}\in\Theta} \sum_{t=s}^{s+\tau-1} f_t(\mathcal{G}_t(\boldsymbol{\theta}))$ where $[s,s+\tau-1]\subseteq[T]$.
    Under Assumptions \ref{assmp1}, \ref{assmp2} and \ref{assmp3}, the regret FairSAR proposed in Eq.(\ref{eq:ourRegret}) of \sysname{} in Algorithm \ref{alg:ouralgorithm} satisfies
    \begin{align*}
    &\max_{[s,s+\tau-1]\subseteq[T]} \bigg( \sum_{t=s}^{s+\tau-1} f_t\Big(\mathcal{G}_t(\boldsymbol{\theta}_t)\Big) - f_t\Big(\mathcal{G}_t(\boldsymbol{\theta}^*)\Big) \bigg) \leq O\Big((\tau\log T)^{1/2}\Big)\\
    &\max_{[s,s+\tau-1]\subseteq[T]} \bigg (\sum_{t=s}^{s+\tau-1} g_i\Big(\mathcal{G}_t(\boldsymbol{\theta}_t)\Big) \bigg)\leq O\Big((\tau T\log T)^{1/4}\Big), \quad \forall i \in[m]
    \end{align*}
\end{sloppypar}
\end{theorem}
\begin{table*}[t]
\caption{Comparison of upper bounds in loss regret and constraint violations across various methods.}
\vspace{-3mm}
\setlength\tabcolsep{3pt}
\begin{tabular}{cccccccc}
\cline{2-8}
\multicolumn{1}{l}{} & \multicolumn{4}{c}{Static Environment} & \multicolumn{3}{|c}{Changing Environment} \\ 
\hline
\multicolumn{1}{c|}{Algorithms}  & \multicolumn{1}{c}{FTML\cite{Finn-ICML-2019}} & \multicolumn{1}{c}{FairFML\cite{zhao-KDD-2021}} & \multicolumn{1}{c}{FairAOGD\cite{AdpOLC-2016-ICML}} & FairGLC\cite{GenOLC-2018-NeurIPS} & \multicolumn{1}{|c}{AOD\cite{zhang-2020-AISTATS}} & \multicolumn{1}{c}{CBCE\cite{Jun-2017-AISTATS}} & FairSAOML(Ours) \\ 
\hline
\multicolumn{1}{c|}{Loss Regret} & \multicolumn{1}{c}{$O(\log T)$} & \multicolumn{1}{c}{$O(\log T)$} & \multicolumn{1}{c}{$O(T^{2/3})$} & $O(\log T)$ & \multicolumn{1}{|c}{$O\big((\tau\log T)^{1/2}\big)$} & \multicolumn{1}{c}{$O\big((\tau\log T)^{1/2}\big)$} & $O\big((\tau\log T)^{1/2}\big)$ \\ 
\hline
\multicolumn{1}{c|}{\begin{tabular}[c]{@{}c@{}} Constraint Violations\end{tabular}} & \multicolumn{1}{c}{-} & \multicolumn{1}{c}{$O\big((T\log T)^{1/2}\big)$} & \multicolumn{1}{c}{$O(T^{2/3})$} & $O\big((T\log T)^{1/2}\big)$ & \multicolumn{1}{|c}{-} & \multicolumn{1}{c}{-} & $O\big((\tau T\log T)^{1/4}\big)$ \\ 
\hline
\end{tabular}
\label{tab:analysis comparison}
\vspace{-4mm}
\end{table*}
\textbf{Discussion for Upper Bounds.} Under aforementioned assumptions and provable convexity of Eq.(\ref{eq:L_t regret}) in $\boldsymbol{\theta}$ (see Lemma \ref{lemma:convex-concave} in Appendix \ref{App:proof}), the proposed \sysname{} in Algorithm \ref{alg:ouralgorithm} achieves sub-linear bounds in FairSAR for both loss regret and violation of fairness constraints. 
Although such bounds are comparable with the strongly adapted loss regret in \cite{Jun-2017-AISTATS,zhang-2020-AISTATS} (see Table \ref{tab:analysis comparison})
in terms of online learning in changing environment paradigms, for the first time we bound loss regret and cumulative fairness constraints simultaneously. On the other hand, in terms of fairness-aware online learning, our proposed method outperforms \cite{zhao-KDD-2021,AdpOLC-2016-ICML,GenOLC-2018-NeurIPS} by giving a tighter bound of fair constraint violations.

\textbf{Complexity.} The computational complexity of \sysname{} in Algorithm \ref{alg:ouralgorithm} at each round $t\in[T]$ is $O(N_{meta}\cdot |\mathcal{U}|)$ where $N_{meta}$ is the number of meta-level iterations and $|\mathcal{U}|=O(\log T)$ is the total number of experts that needs to be maintained, and the complexity of each expert is $O(1)$.

\section{Experimental Settings}
\label{sec:experiments}


\textbf{Datasets.} We use the following publicly available datasets. 
(1) \textit{New York Stop-and-Frisk} \cite{Koh-icml-2021} is a prominent dataset of a real-world application on policing in the new york city from 2009 to 2010. It documents whether a pedestrian who was stopped on suspicion of weapon possession would in fact possess a weapon. 
As this data had a pronounced racial bias on African Americans, for each frisked record, we consider race as the binary protected attribute, that is black and non-black. Besides, this dataset consists of records collected in five different sub-districts, Manhattan (\textbf{M}), Brooklyn (\textbf{B}), Queens (\textbf{Q}), Bronx (\textbf{R}) and Staten (\textbf{S}). Since there are large performance disparities across districts and race groups, each district is viewed as an independent domain. To adapt the online learning setting, data in each domain is further split into 32 tasks and each task corresponds to ten days of a month with 111 non-protected features. According to \textit{DDP} values in Definition \ref{dbc definition}, the fairness levels from low to high are Bronx (0.74), Queens (0.68), Staten (0.65), Manhattan (0.53) and Brooklyn (0.44). The larger \textit{DDP} values indicate lower fairness level.
We hence consider two settings for domain adaptation where each setting contains 96 tasks in total: (i) fairness level from high to low: Brooklyn to Manhattan to Staten (\textbf{B}$\rightarrow$\textbf{M}$\rightarrow$\textbf{S}); and (ii) fairness level from low to high: Bronx to Queens then Staten (\textbf{R}$\rightarrow$\textbf{Q}$\rightarrow$\textbf{S}).
(2) \textit{MovieLens}\footnote{https://grouplens.org/datasets/movielens/100k/} contains 100k ratings by 943 users on 1682 movies and each rating is given a binary label ("recommending" if rating greater than 3, ``not recommending" otherwise). We consider gender as the protected attribute. To generate dynamic environments, following \cite{Wan_2021_AAAI} we construct a larger dataset by combining three copies of the original data and flipping the original values of non-protected attributes by multiplying -1 for the middle copy. Therefore, each copy is considered as a data domain. Furthermore, each data copy is split into 30 tasks by timestamps and there are 90 tasks in total.

\textbf{Evaluation Metrics.} Two popular evaluation metrics are introduced that each allows quantifying the extent of bias taking into account the protected attribute.
\textit{Demographic Parity} (DP) \cite{Dwork-2011-CoRR} and \textit{Equalized Odds} (EO) \cite{Hardt-NIPS-2016} can be formalized as
\begin{align*}
    \text{DP} = \frac{P(\hat{Y}=1|S=0)}{P(\hat{Y}=1|S=1)}; \quad \text{EO} = \frac{P(\hat{Y}=1|S=0,Y=y)}{P(\hat{Y}=1|S=1,Y=y)}
\end{align*}
where $y\in\{-1,1\}$. Equalized odds requires that $\hat{Y}$ have equal true positive rates and false positive rates between sub-groups. For both metrics, a value closer to 1 indicate fairness.

\textbf{Competing Methods.} We compare the performance of our algorithm \sysname{} with six baseline methods. These baselines are chosen from three perspectives: online meta learning (MaskFTML, FairFML), online fairness learning (FairFML, FairAOGD, FairGLC) and online learning in changing environment (AOD, CBCE).
(1) \textbf{MaskFTML} \cite{Finn-ICML-2019}: the original FTML finds a sequence of meta parameters by simply applying MAML \cite{Finn-ICML-2017-(MAML)} at each round. To focus on fairness learning, this approach is applied to modified datasets by simply removing protected attributes. 
(2) \textbf{FairFML} \cite{zhao-KDD-2021} controls bias in an online working paradigm and aims to attain zero-shot generalization with task-specific adaptation. Different from our \sysname{}, FairFML focuses on static environment and assumes tasks sampled from an unchangeable distribution.
(3) \textbf{FairAOGD} \cite{AdpOLC-2016-ICML} is proposed for online learning with long-term constraints. In order to fit bias-prevention and compare them to \sysname{}, we specify such constraints as DDP stated in Definition \ref{dbc definition}.
(4) \textbf{FairGLC} \cite{GenOLC-2018-NeurIPS} rectifies FairAOGD by square-clipping the constraints in place of $g_i(\cdot), \forall i$.
(5) \textbf{AOD} \cite{zhang-2020-AISTATS} minimizes the strongly adaptive regret by running multiple online gradient descent algorithms over a set of dense geometric covering intervals. 
(6) \textbf{CBCE} \cite{Jun-2017-AISTATS} adapts changing environment in an online learning paradigm by combining the sleeping bandits idea with the coin betting algorithm.

\begin{figure*}[!h]
\captionsetup[subfigure]{aboveskip=-1pt,belowskip=-1pt}
\centering

    \begin{subfigure}[b]{0.245\textwidth}
        \includegraphics[width=\textwidth]{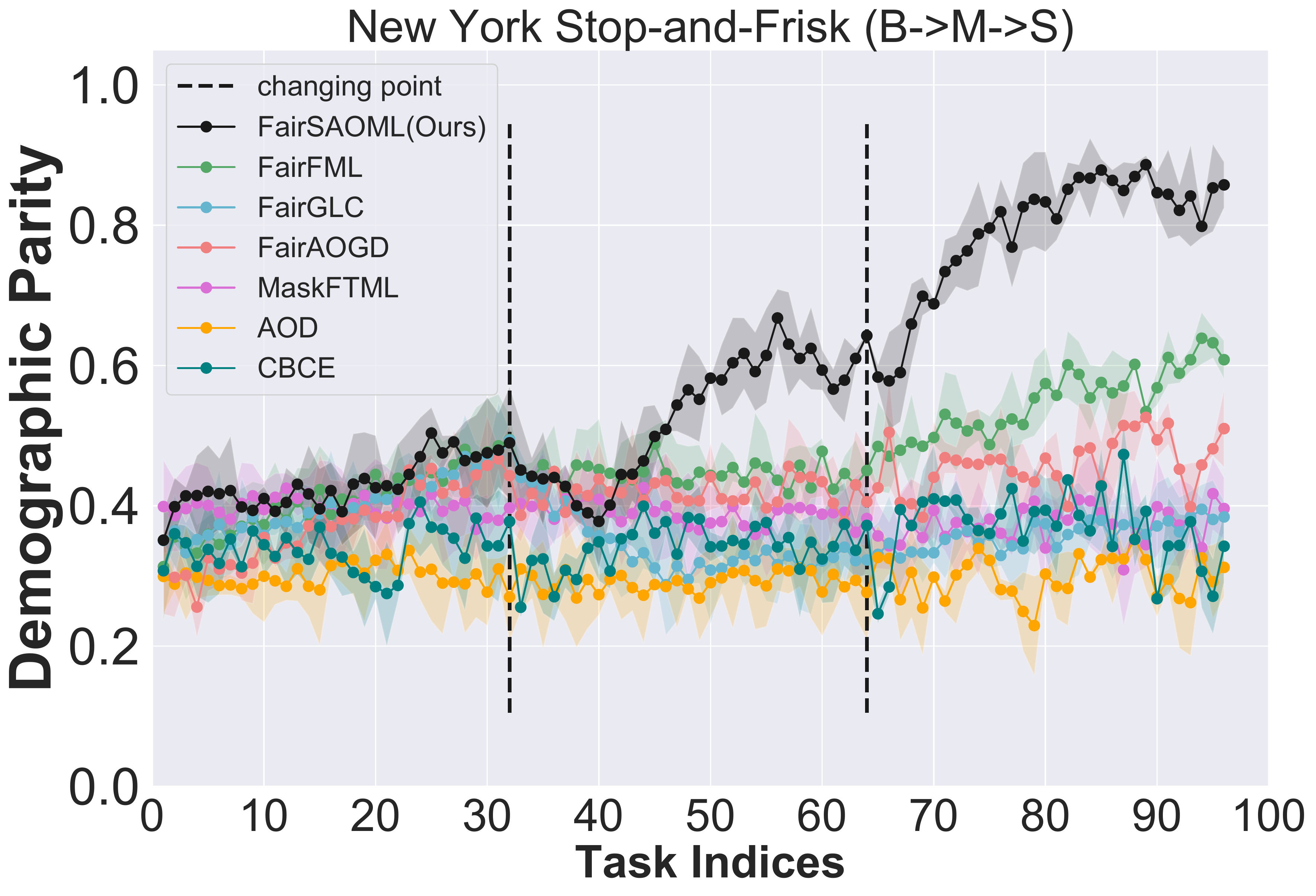}
        \caption{}
    \end{subfigure}
    \begin{subfigure}[b]{0.245\textwidth}
        \includegraphics[width=\textwidth]{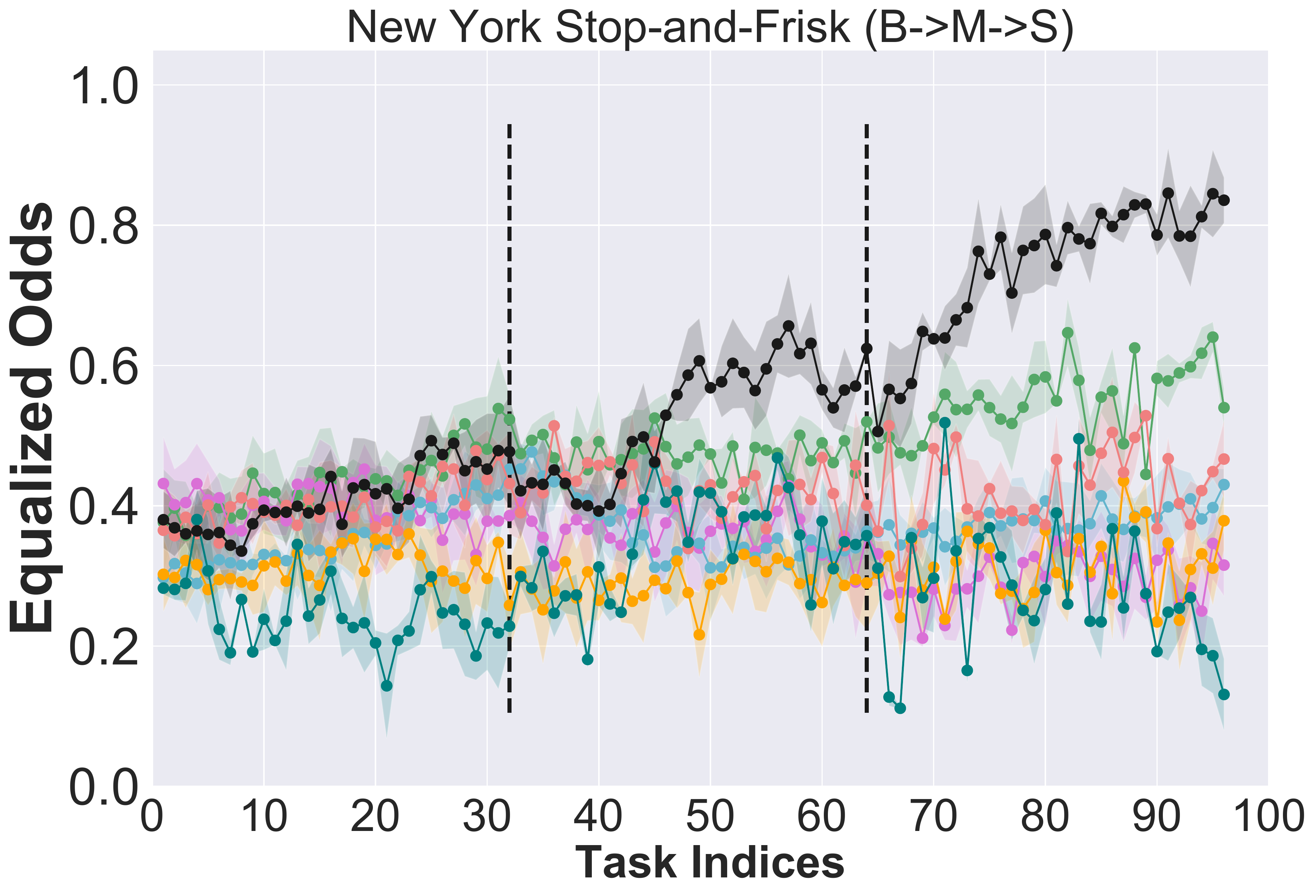}
        \caption{}
    \end{subfigure}
    \begin{subfigure}[b]{0.245\textwidth}
        \includegraphics[width=\textwidth]{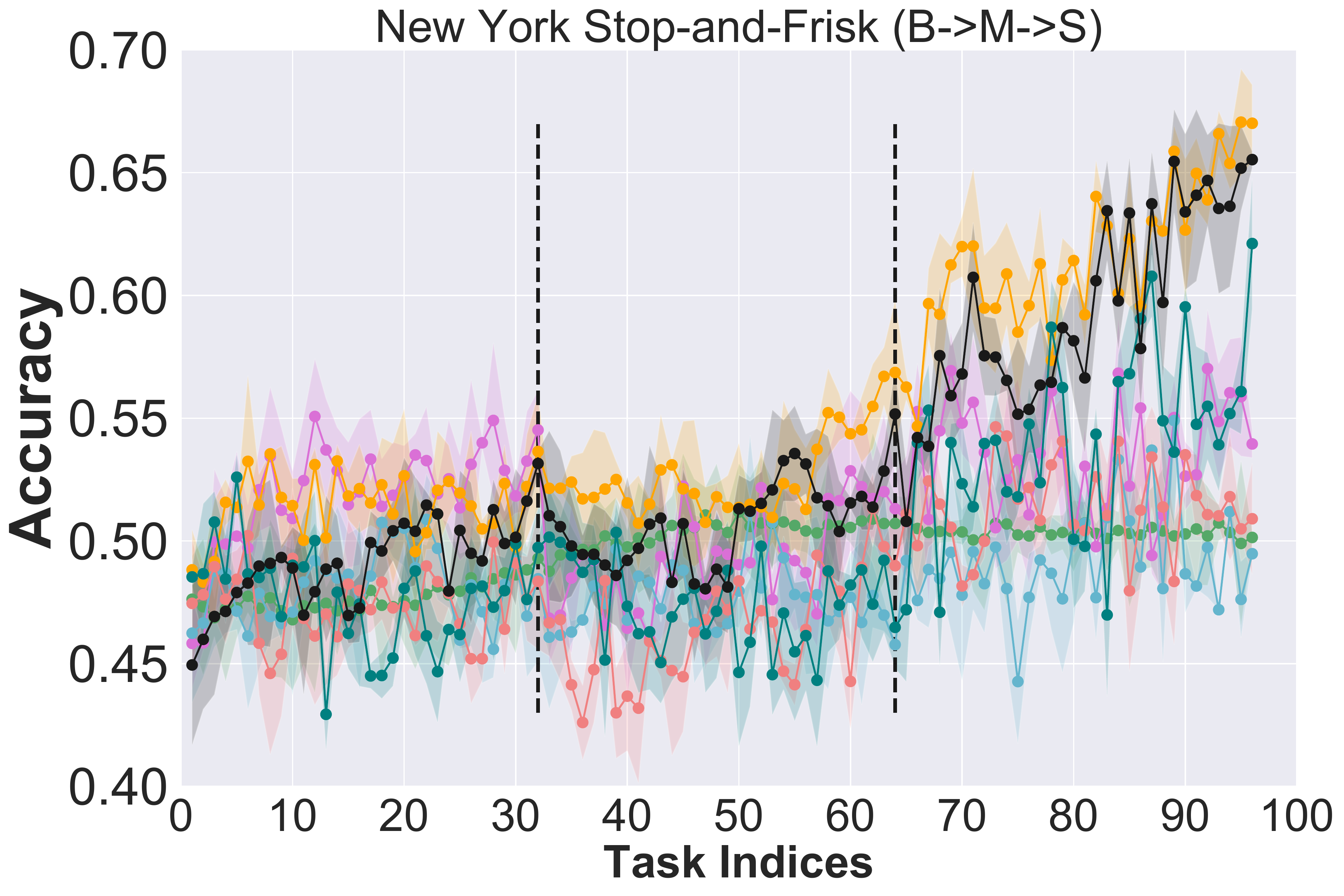}
        \caption{}
    \end{subfigure}
    \begin{subfigure}[b]{0.245\textwidth}
        \includegraphics[width=\textwidth]{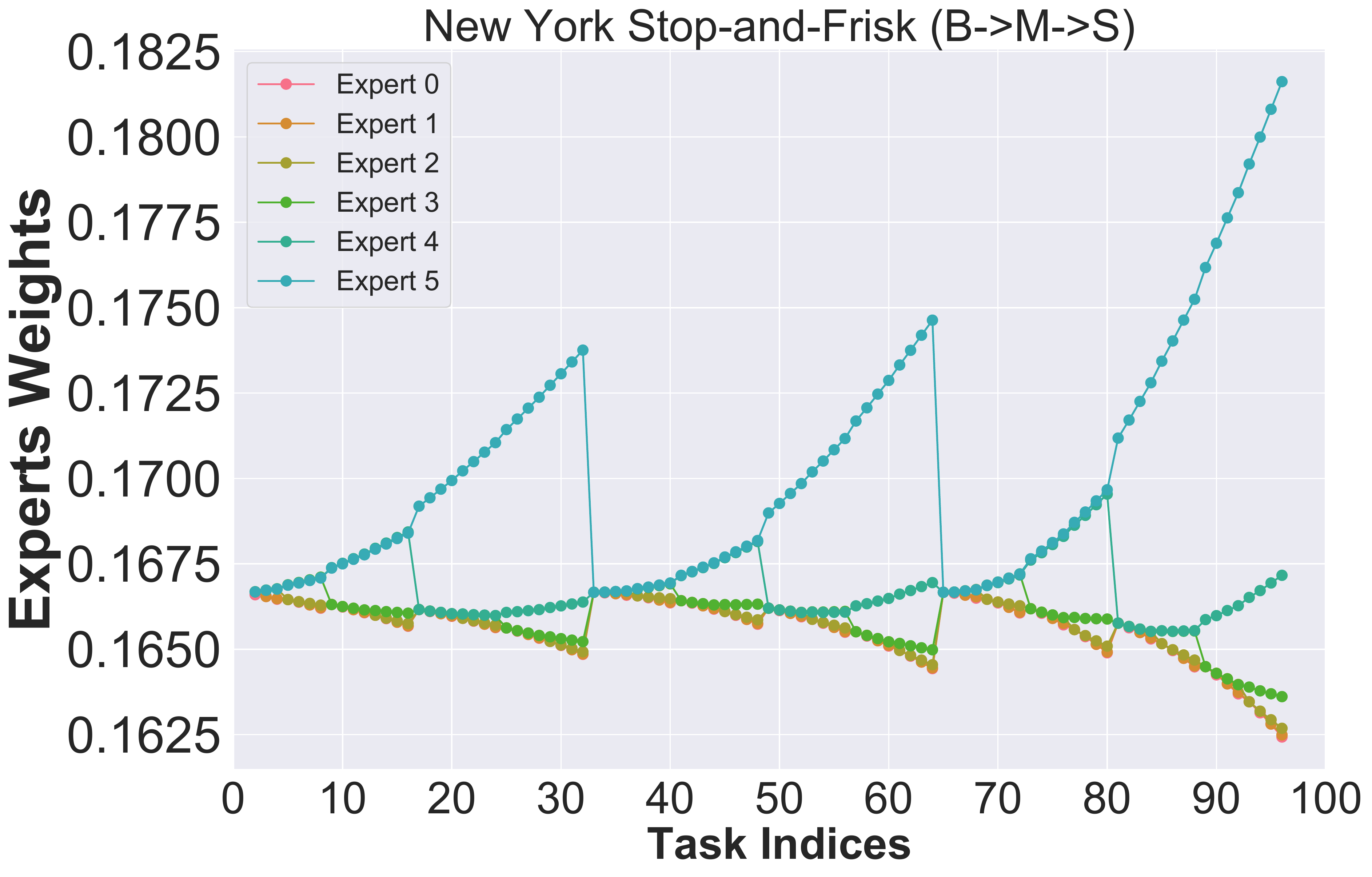}
        \caption{}
    \end{subfigure}

    \begin{subfigure}[b]{0.245\textwidth}
        \includegraphics[width=\textwidth]{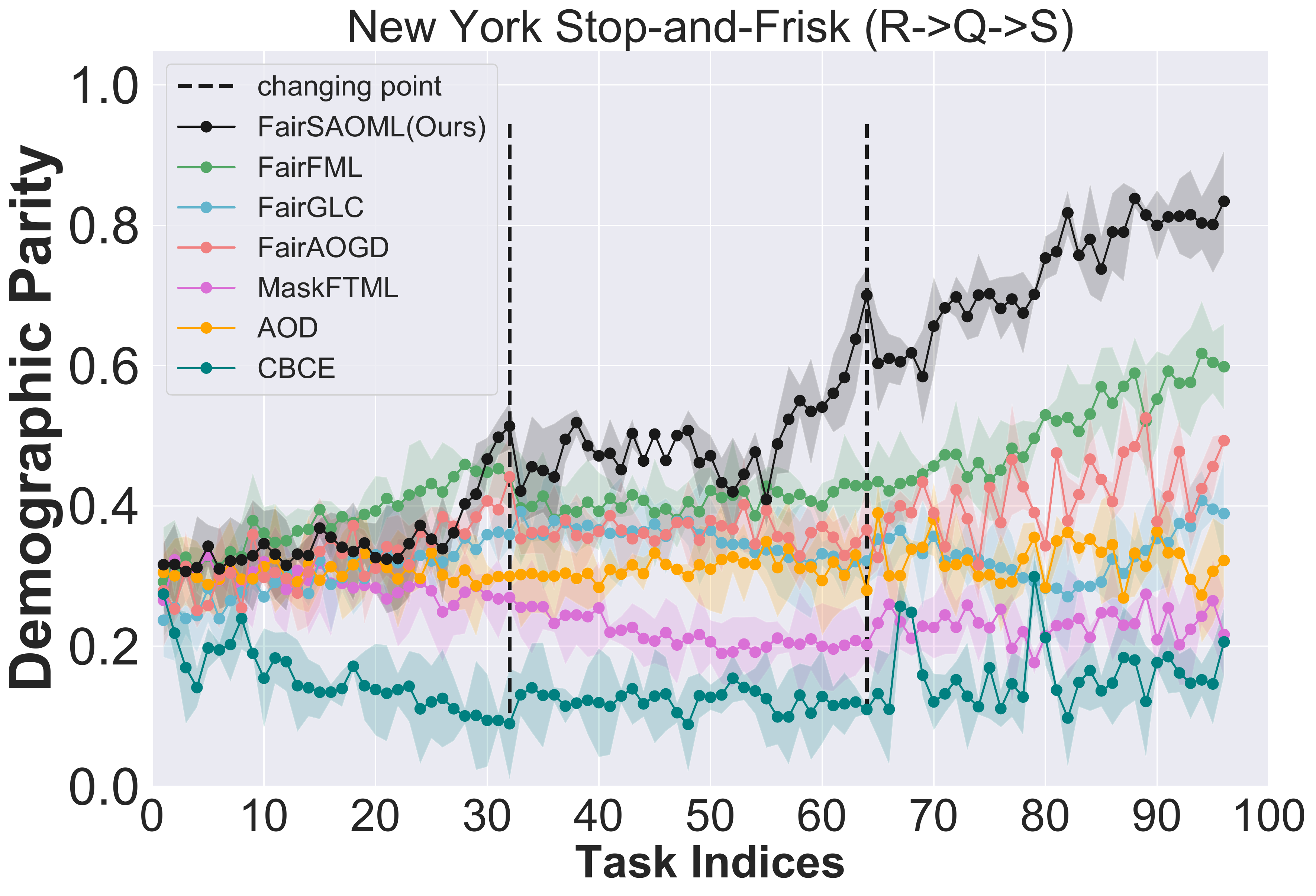}
        \caption{}
    \end{subfigure}
    \begin{subfigure}[b]{0.245\textwidth}
        \includegraphics[width=\textwidth]{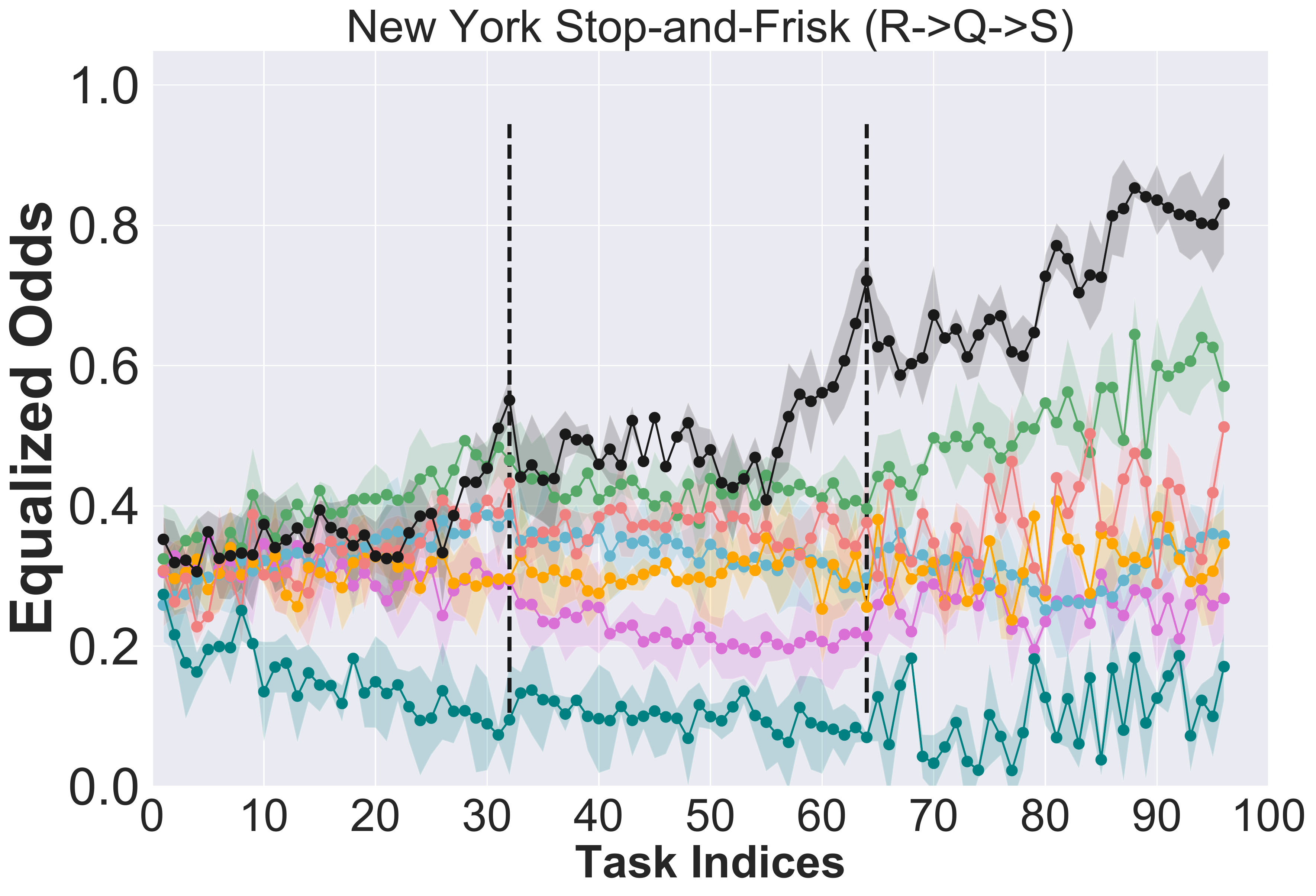}
        \caption{}
    \end{subfigure}
    \begin{subfigure}[b]{0.245\textwidth}
        \includegraphics[width=\textwidth]{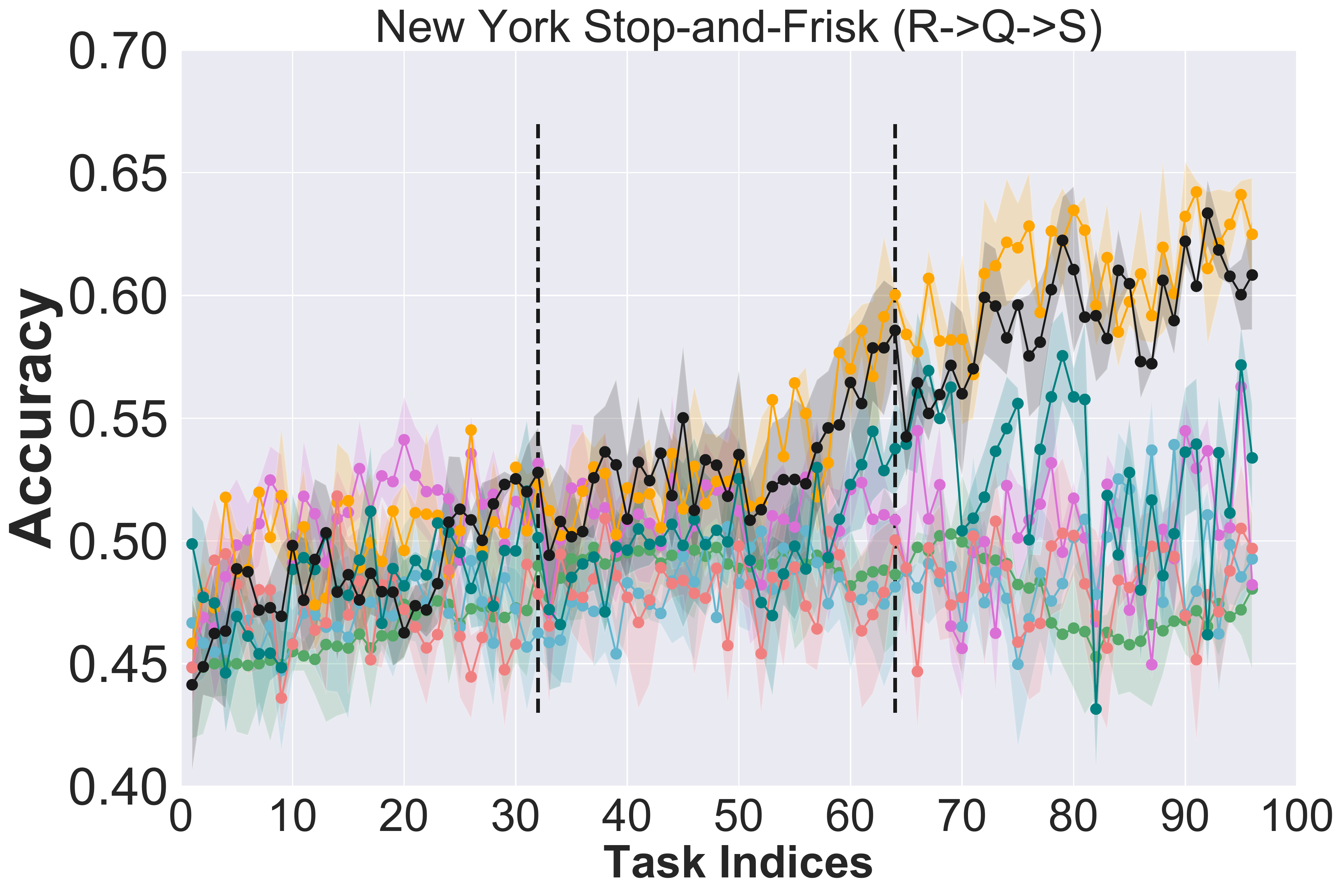}
        \caption{}
    \end{subfigure}
    \begin{subfigure}[b]{0.245\textwidth}
        \includegraphics[width=\textwidth]{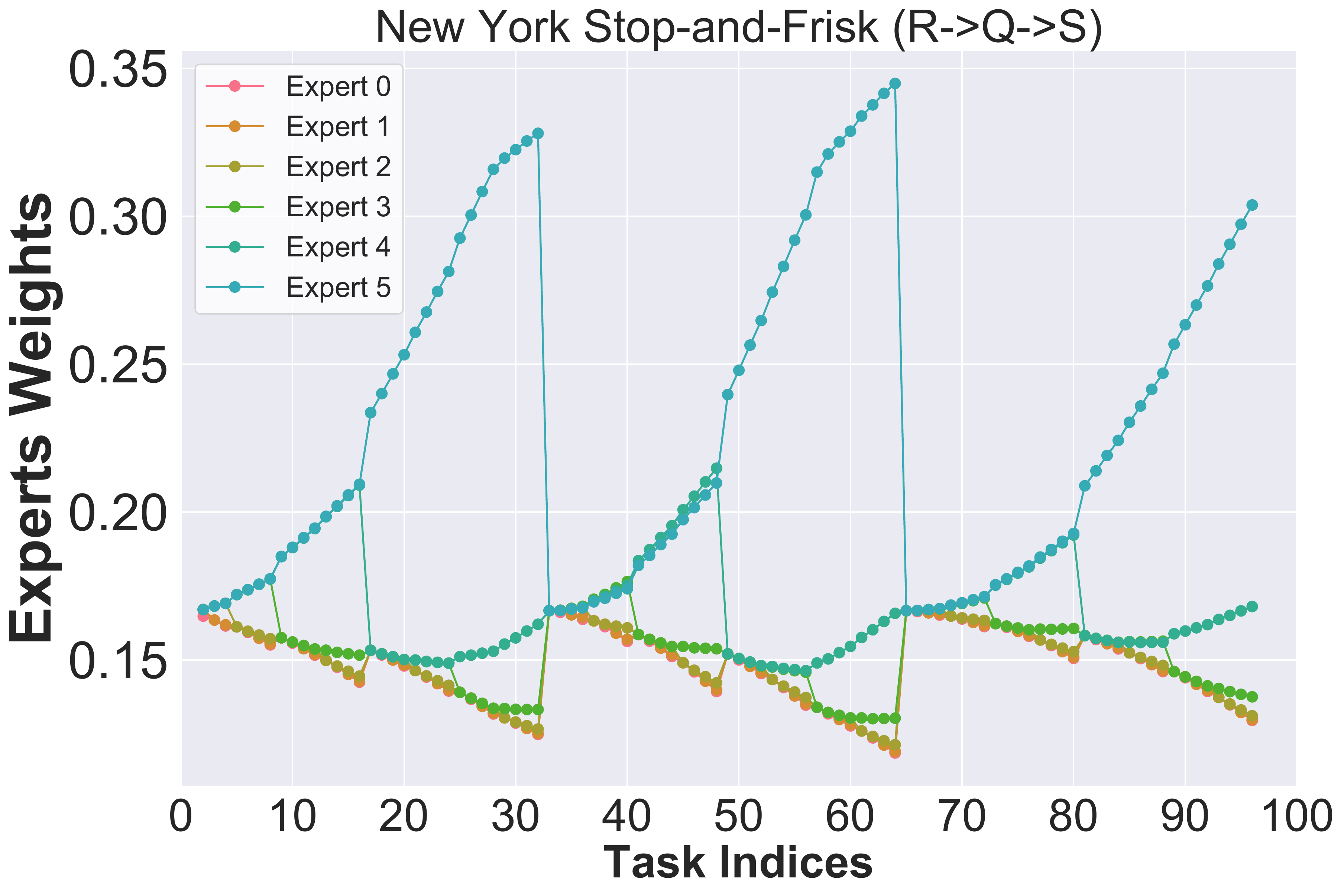}
        \caption{}
    \end{subfigure}

    \begin{subfigure}[b]{0.245\textwidth}
        \includegraphics[width=\textwidth]{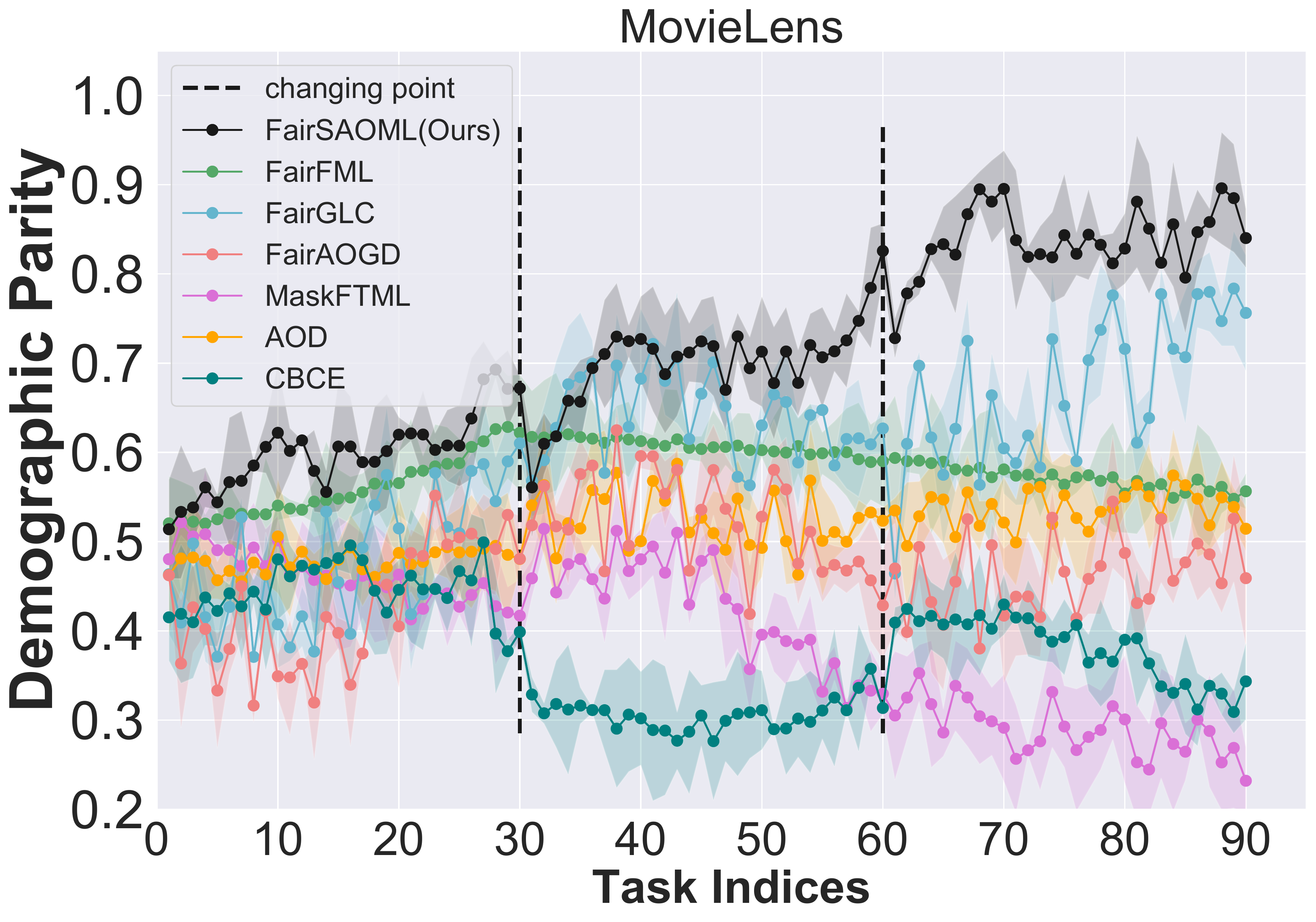}
        \caption{}
    \end{subfigure}
    \begin{subfigure}[b]{0.245\textwidth}
        \includegraphics[width=\textwidth]{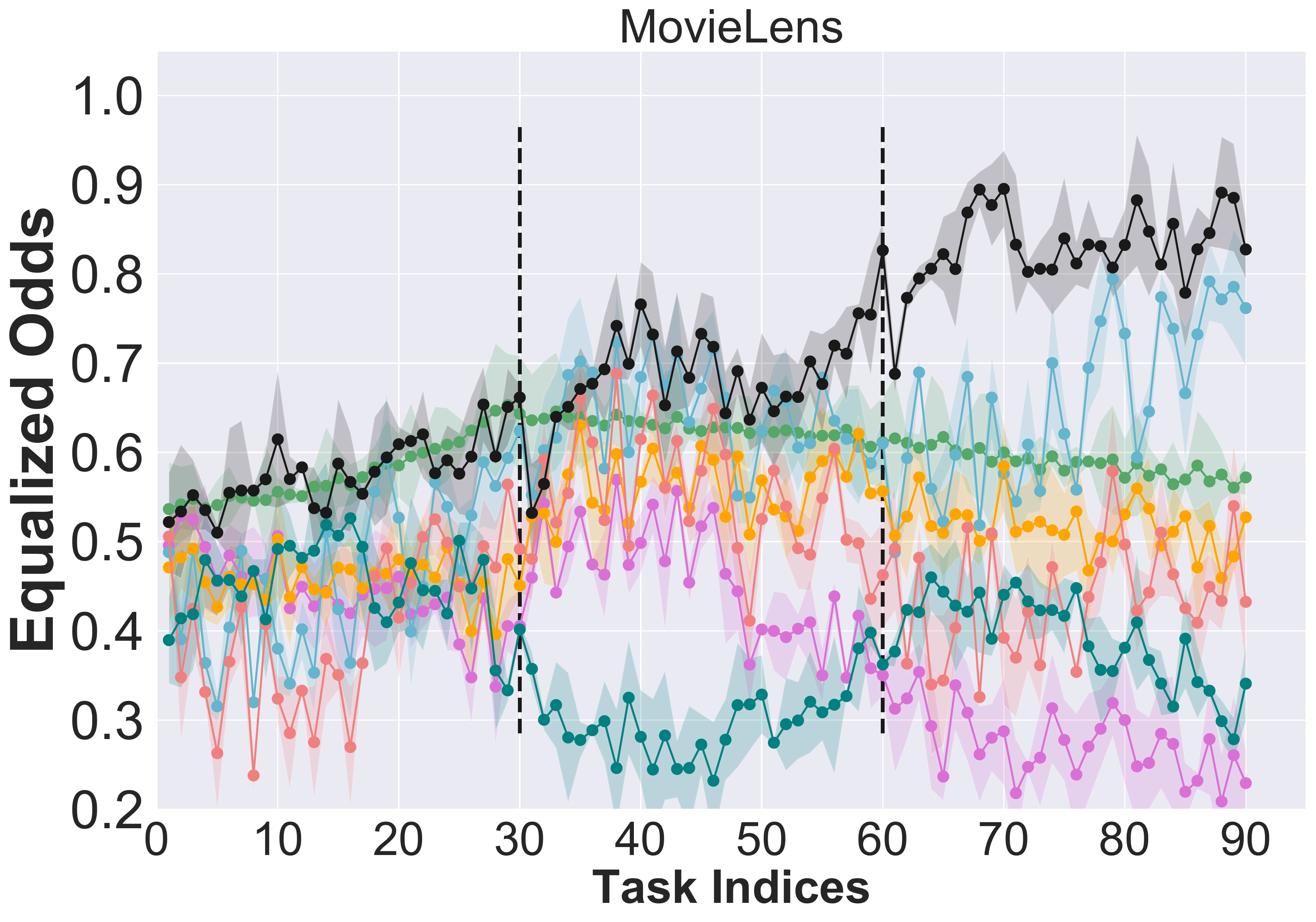}
        \caption{}
    \end{subfigure}
    \begin{subfigure}[b]{0.245\textwidth}
        \includegraphics[width=\textwidth]{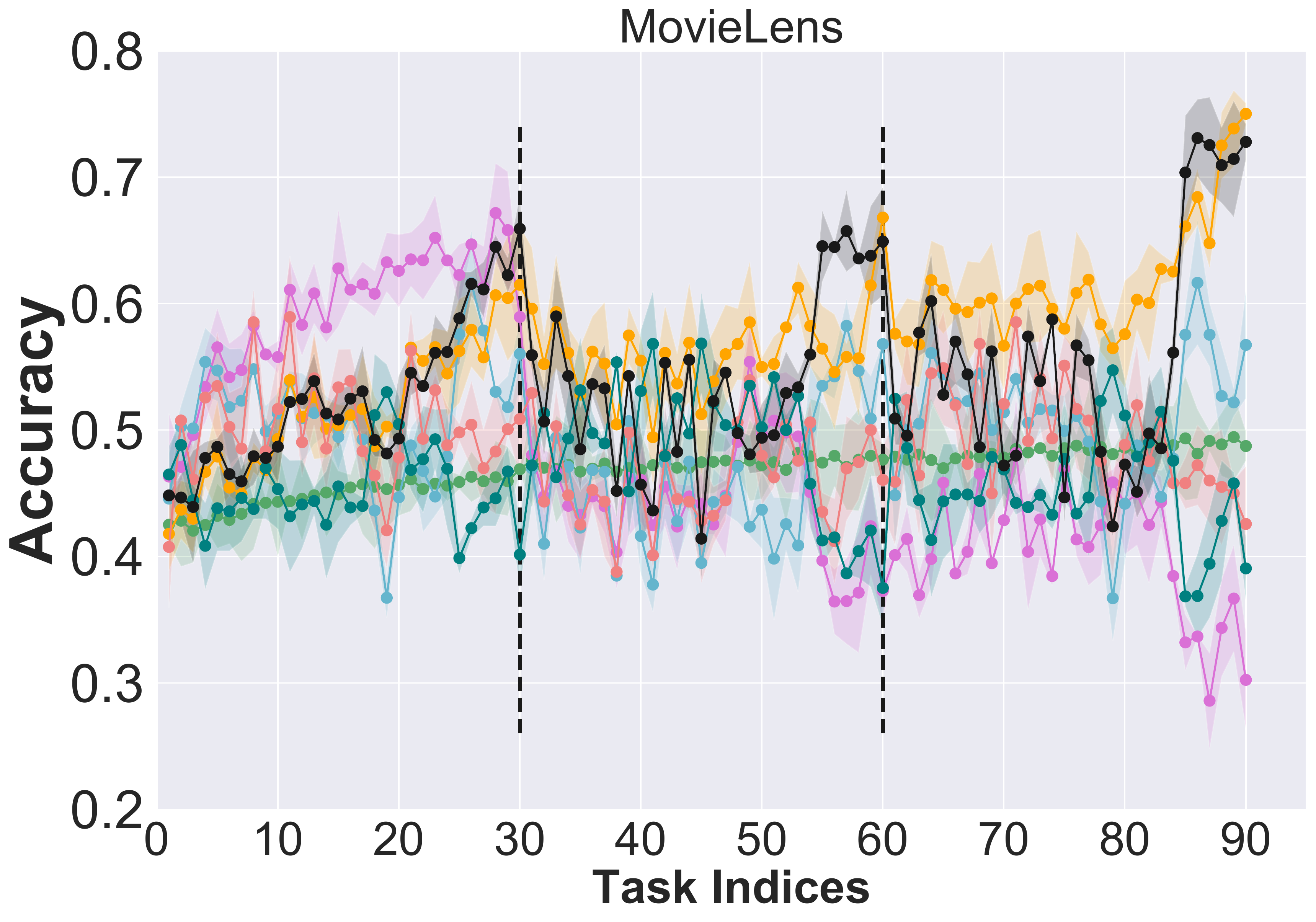}
        \caption{}
    \end{subfigure}
    \begin{subfigure}[b]{0.245\textwidth}
        \includegraphics[width=\textwidth]{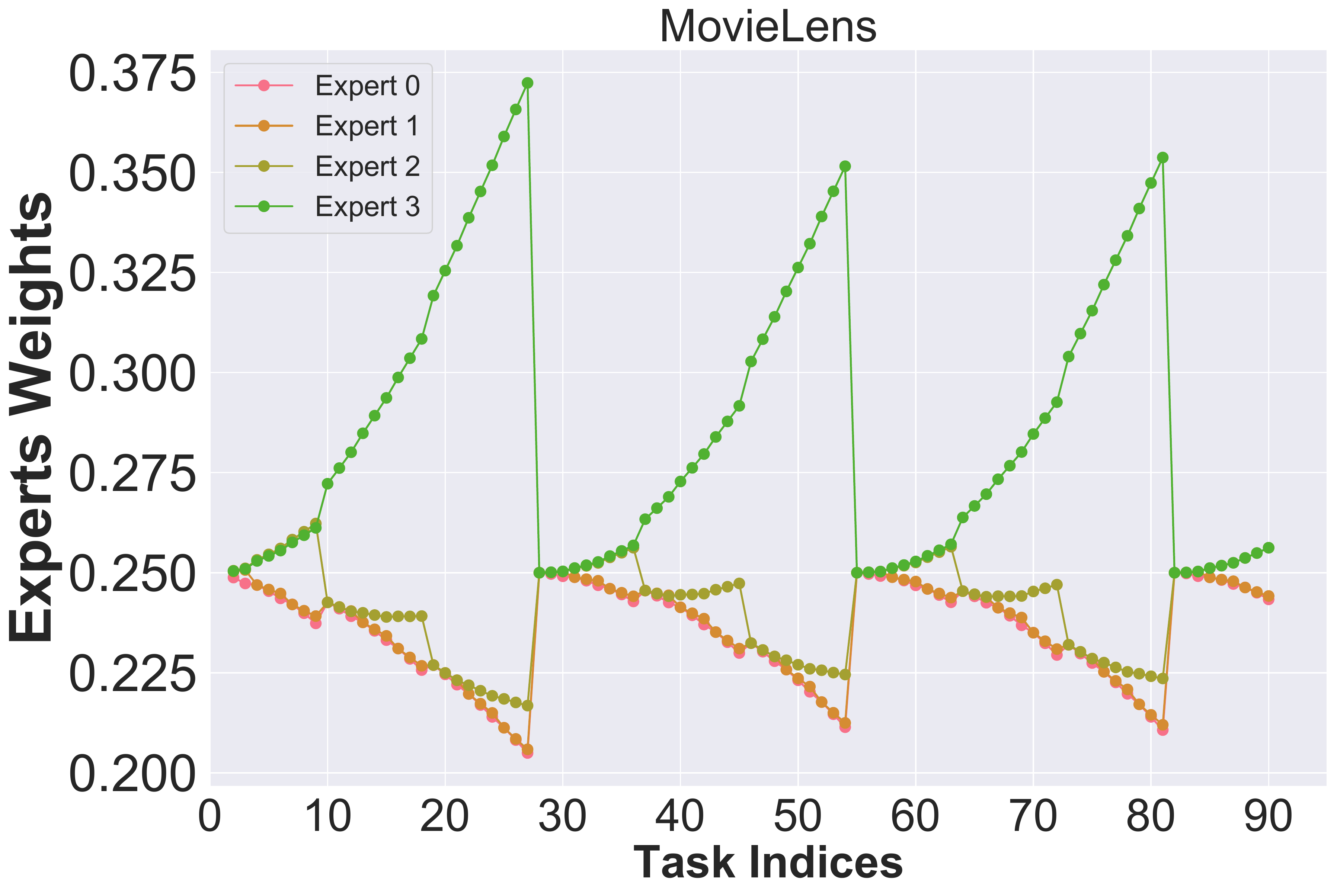}
        \caption{}
    \end{subfigure}
    \vspace{-3mm}
    \caption{Model performance over real-world datasets through each round. New York Stop-and-Frisk (a-d) \textbf{B}$\rightarrow$\textbf{M}$\rightarrow$\textbf{S}, (e-h) \textbf{R}$\rightarrow$\textbf{Q}$\rightarrow$\textbf{S}; (i-l) MovieLens.}
    \label{fig:dp-eop-acc}
\vspace{-3mm}
\end{figure*}


\textbf{Settings.} As discussed in Sec.\ref{sec:analysis}, the performance of our proposed method has been well justified theoretically for machine learning models
whose objectives are strongly convex and smooth.
However, in machine learning and fairness studies, due to the non-linearity of neural networks, many problems have a non-convex landscape where theoretical analysis is challenging. Nevertheless, algorithms originally developed for convex optimization problems like gradient descent have shown promising results in practical non-convex settings \cite{Finn-ICML-2019}. Taking inspiration from these successes, we describe practical instantiations for the proposed online algorithm, and empirically evaluate the performance in Sec.\ref{sec:results}. 

For each task we set the number of fairness constraints to one, \textit{i.e.} $m=1$. For the rest, we follow the same settings as used in online meta learning \cite{Finn-ICML-2019,zhao-KDD-2021}. In particular, we meta-train with support size of 400 for each class and 800 for a query set, whereas $90\%$ (hundreds of datapoints) of task samples for evaluation. 
Besides, for \textit{New York Stop-and-Frisk} dataset, we choose base $=2$ and the total number of experts is $\lfloor\log_2^{96}\rfloor=6$. Similarly, we choose base $=3$ for the \textit{MovieLens} dataset and hence it has $\lfloor\log_3^{90}\rfloor=4$ experts in total.
All the baseline models that are used to compare with our proposed approach share the same neural network architecture and parameter settings.
All the experiments are repeated 10 times with the same settings and the mean and standard deviation results are reported. 
Details on the settings and hyperparameter tuning are given in Appendix \ref{app:Data Pre-Processing} and \ref{app:Implementation Details and Hyperparameter Tuning}.

\section{Results}
\label{sec:results}
    \begin{figure*}
    \begin{minipage}{0.25\textwidth}
        \centering
        \includegraphics[width=\textwidth]{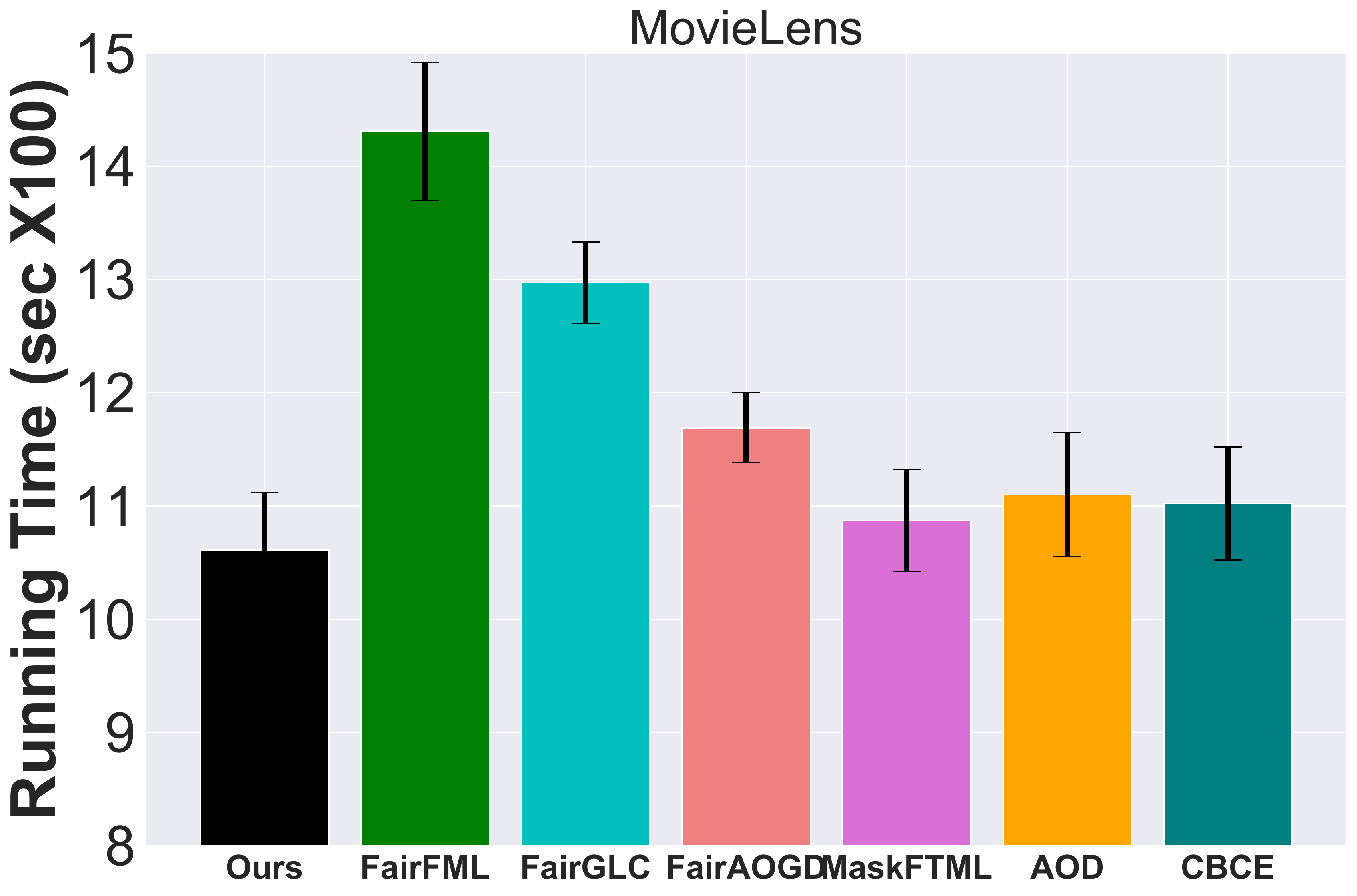}
        \vspace{-7mm}
        \caption{Running time comparison over baseline methods}
        \label{fig:efficiency}
        \vspace{-3mm}
    \end{minipage}\hfill
    \begin{minipage}{0.75\textwidth}
        \centering
        \begin{subfigure}[b]{0.325\textwidth}
            \includegraphics[width=\textwidth]{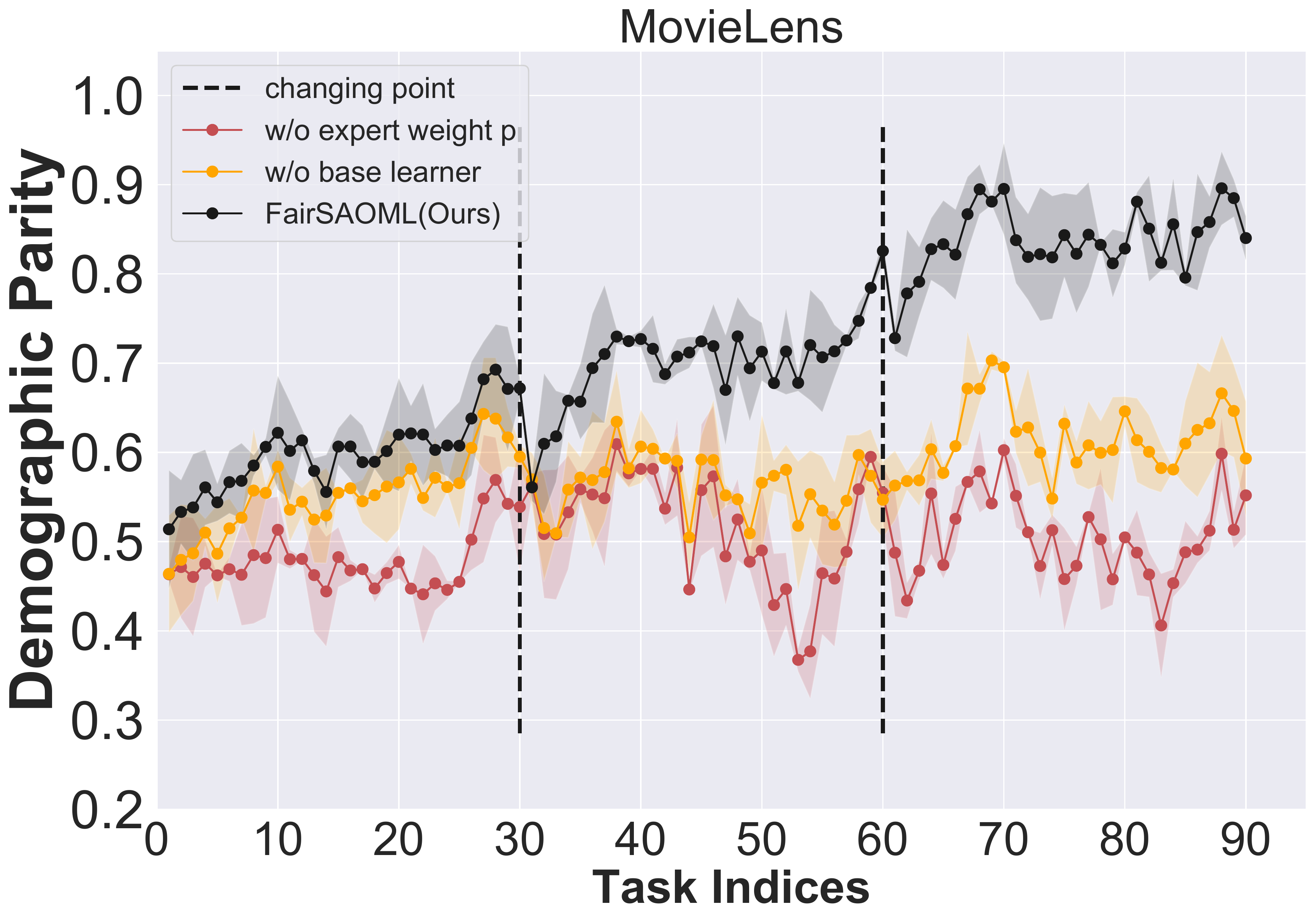}
        \end{subfigure}
        \begin{subfigure}[b]{0.325\textwidth}
            \includegraphics[width=\textwidth]{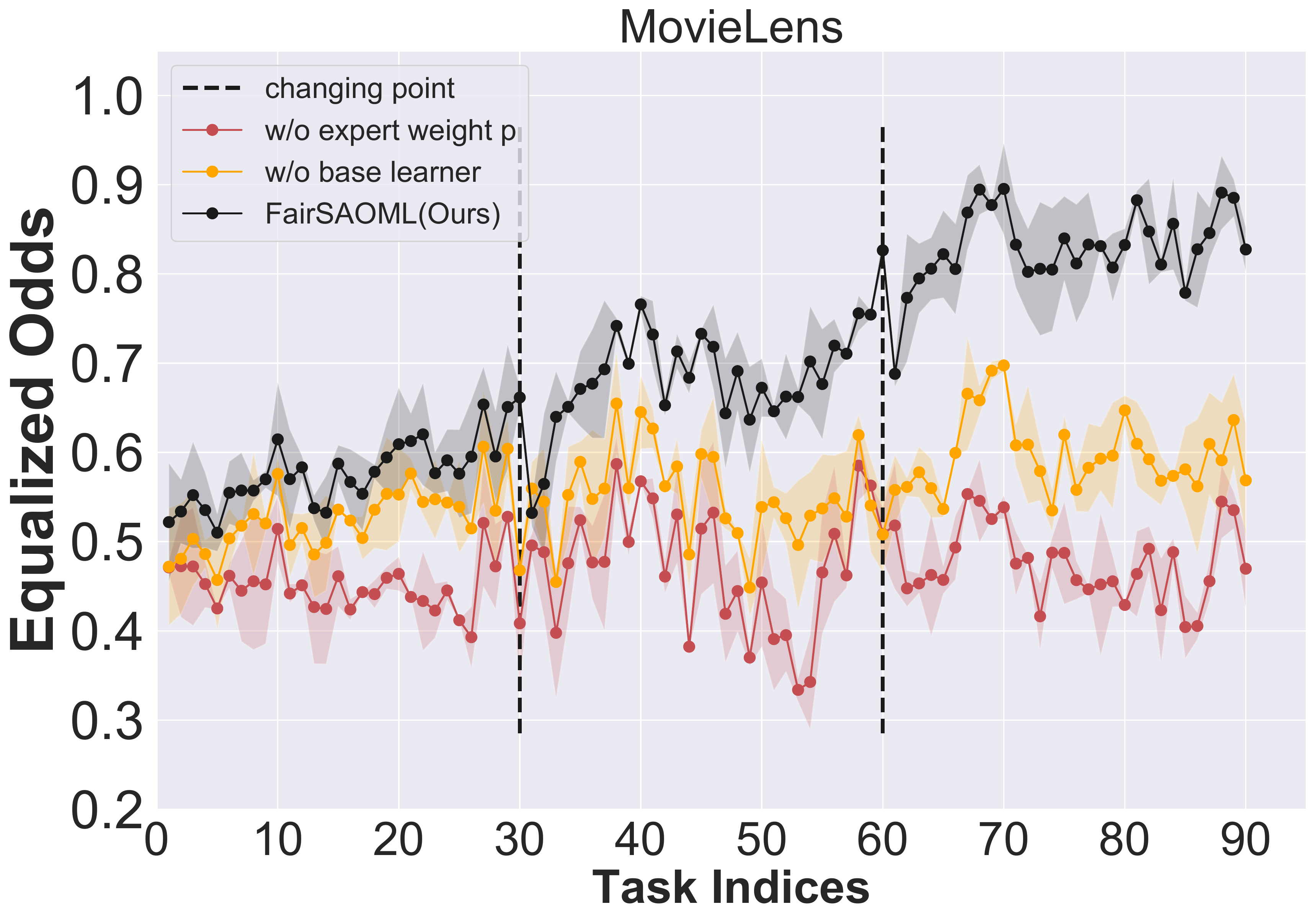}
        \end{subfigure}
        \begin{subfigure}[b]{0.325\textwidth}
            \includegraphics[width=\textwidth]{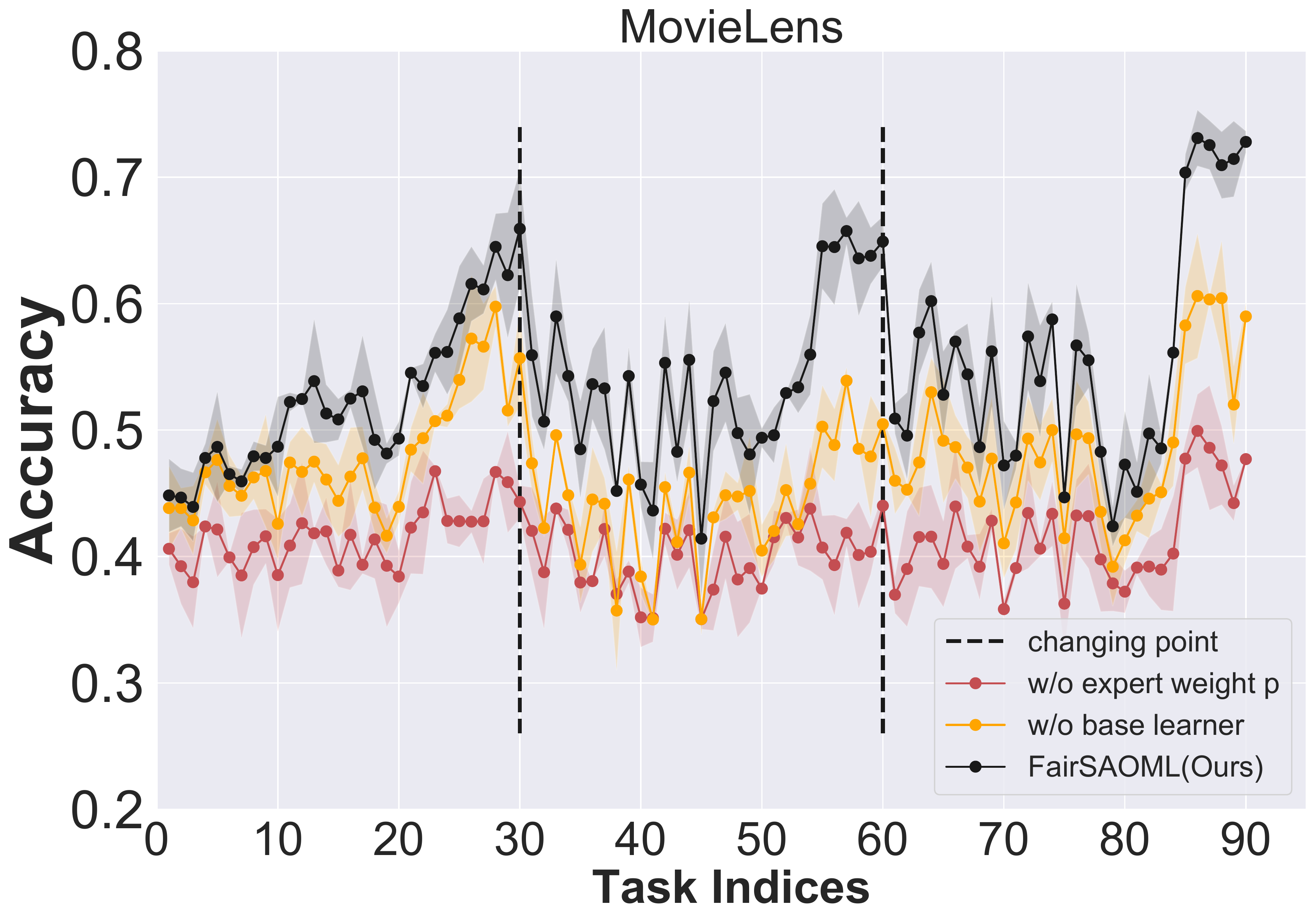}
        \end{subfigure}
        \vspace{-3mm}
        \caption{Ablation studies on the MovieLens dataset.}
        \label{fig:ablation}
        \vspace{-3mm}
    \end{minipage}
\end{figure*}

\subsection{Overall Performance}

As for all methods showcased in in Figure \ref{fig:dp-eop-acc}, higher is better for all plots, while shaded regions indicate standard errors. The learning curves show that the proposed \sysname{} effectively controls bias as the learner sees more tasks, and it eventually meets the fair condition of "80\%-rule" \cite{Biddle-Gower-2005} where DP and EO at last several rounds are beyond 0.8.
Furthermore, \sysname{} substantially outperforms most of the alternative approaches in achieving the best model precision represented by high accuracies.
Besides, in terms of learning efficiency, our \sysname{} shows the least running time compared with baselines in Figure \ref{fig:efficiency}. This results from that (1) at each time the total number of experts are fixed (\textit{i.e.} 4 in \textit{MovieLens}), and (2) instead of using the entire data task, only a subset (support) of data is used for parameters update of active experts.

\vspace{-5mm}
\subsection{Adaptability to Changing Environments}
The main focus of the designed experiments is to test the adaptability of our \sysname{} with respect to both fairness and model accuracy from one data domain to another.
To better visualize changing environments, we manually add vertical dotted lines in Figure \ref{fig:dp-eop-acc} to distinguish different domains at task indices of changing data domains.
Our experimental results demonstrate that although \sysname{} may not dominate the performance over other baseline methods in the first data domain, it can effectively adapt to changes in the environment and so that the performance is constantly improving in both bias control and predictive accuracy. 

In Sec. \ref{sec:agc intervals and experts}, we introduce that experts serve as key components in \sysname{} and model parameters pair $(\boldsymbol{\theta}_t,\boldsymbol{\lambda}_t)$ at round $t$ are determined by combining weighted expert advises (see Figure \ref{fig:overview}). Figure \ref{fig:dp-eop-acc} (d,h,l) show weights change of all experts in \sysname{}. We observe that (1) expert-weight changes periodically; (2) experts associated with longer intervals are assigned with larger weights and these weights keep growing as the learner sees more tasks; (3) on the contrary, smaller weights are given to experts which carry short intervals and become less valued. 
No surprisingly, heavier weights on experts with long intervals enable our \sysname{} to adapt to the instability of model performance caused by changing environments.

As for baseline methods, MaskFTML outperforms other methods in accuracy in the first domain (see Figure \ref{fig:dp-eop-acc} (c,g,k)), but it is not comparable with respect to model fairness. This indicates that attempting to make decision-makers blind to the protected attribute is not able to ameliorate prediction fairness. Although FairFML, FairAOGD, and FairGLC are able to control bias in the first domain, they fail to adapt fairness or predictive accuracy when data domain changes.
Since AOD and CBCE are initially designed for online learning in changing environments, they merely focus on learning accuracy but ignore control model fairness when data domain shifted. Besides, higher accuracy in AOD leads to a trade-off performance on fairness.

\vspace{-3mm}
\subsection{Ablation Studies and Sensitive Analysis}
We conduct ablation studies on the \textit{MovieLens} dataset to demonstrate the contributions of two key components in \sysname{}: expert weights $p_{t,I}$ and the base learner highlighted in Figure \ref{fig:overview}. Specifically, at each round, meta-level parameters are derived by combining weighted expert decisions. By removing expert weights, all experts will contribute equally. Furthermore, in active experts, base learners stated in Eq.(\ref{eq:inner-problem}) are used to further update model parameters at an interval level. Without base learners, all active experts share the same model parameters inherited from the previous round, and hence they are equally weighted. The key findings in Figure \ref{fig:ablation} are (1) expert weights play an important role in \sysname{}, and (2) base learners effectively enhance model performance on bias control and predictive accuracy.

\begin{figure}[!t]
\centering
    \begin{subfigure}[b]{0.235\textwidth}
        \includegraphics[width=\textwidth]{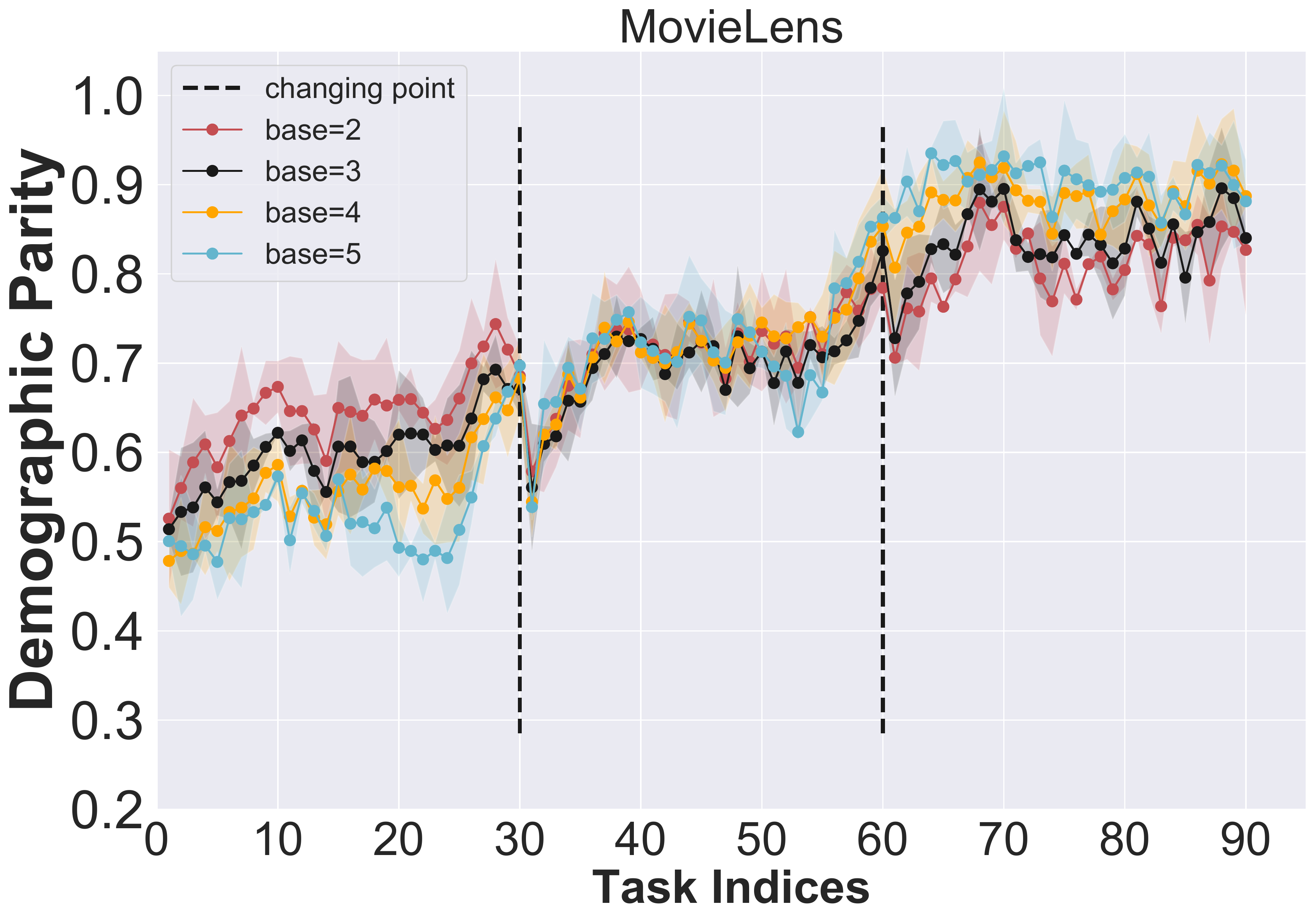}
    \end{subfigure}
    \begin{subfigure}[b]{0.235\textwidth}
        \includegraphics[width=\textwidth]{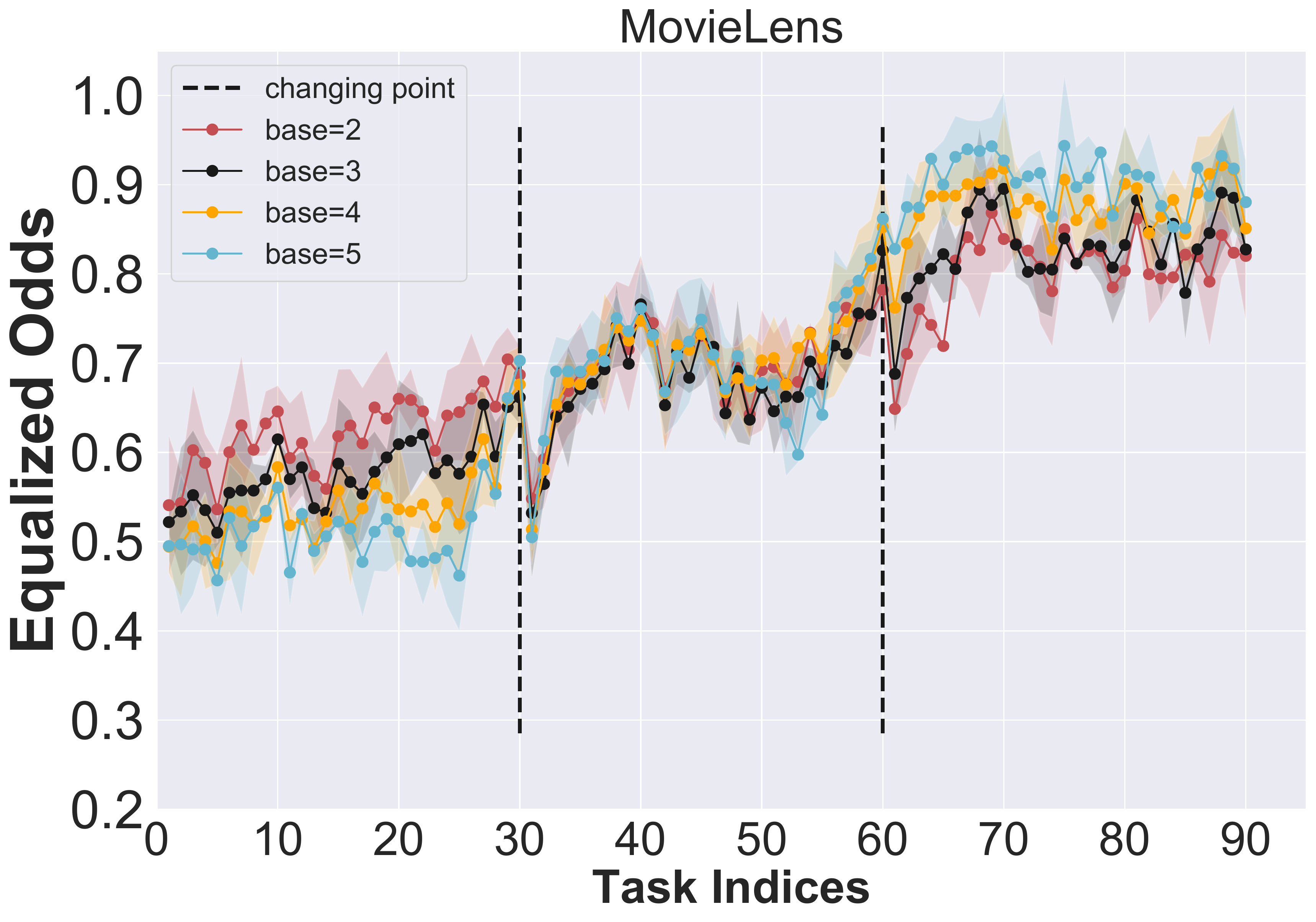}
    \end{subfigure}
    \vspace{-7mm}
    \caption{Sensitive analysis on the MovieLens dataset.}
    \label{fig:base-sensitive}
\end{figure}

Some sensitive analysis on the \textit{MovieLens} dataset are shown in Figure \ref{fig:base-sensitive} where intervals are considered using different bases switched from 2 to 5. According to Eq.(\ref{eq:AGC}), the setting with the smallest base (\textit{i.e.} 2) indicates the most experts (\textit{i.e.} 6) and hence its largest expert carries the longest intervals (\textit{i.e.} 32). We observe that in terms of model fairness, settings with smaller base slightly outperform the ones with larger base in the first domain, but opposite findings are given in the last domain. This is because (1) in the first domain, the largest experts carry more information in smaller base setting than larger base setting; (2) while in the last domain, the largest experts in smaller base settings become impure that take data across two domains. 
More results are given in Appendix \ref{sec:additional result}.

\section{Conclusion}
\label{sec:conclusion}
    To address fairness-aware online learning in changing environments where the learner is faced with data tasks sampled from heterogeneous distributions one after another, we first introduce a novel regret FairSAR which extends strongly adaptive regret by adding fairness-aware long-term constraints. We claim that for the first time a fairness-aware online meta-learning in changing environments framework is proposed.
Then, to determine model parameters at each round, we propose a novel learning algorithm, namely \sysname{}. In this algorithm, we dynamically activate a subset of experts at each round and update their parameters at an interval-level. Model parameters at the meta-level are further achieved by combining weighted actions of all experts. 
Detailed theoretic analysis and corresponding proofs justify the efficiency and effectiveness of the proposed algorithm by demonstrating upper bounds for the loss regret and violation of fair constraints. Empirical studies based on real-world datasets show that our method outperforms state-of-the-art online learning techniques in both model accuracy and fairness.


\begin{acks}
\begin{sloppypar}
The research reported herein was supported by the National Science Foundation (NSF) under grant number OAC-1931541, 2147375, 2137335, IIS-1954376, IIS-1815696, Army Research Office Contract W911NF2110032, and IBM faculty award (Research).
\end{sloppypar}
\end{acks}

\vspace{-3mm}
\bibliographystyle{ACM-Reference-Format}
\bibliography{references}

\newpage
\newpage
\appendix
\section{Additional Experiment Details}
\label{App:ExpDetails}
    \subsection{Notations}
\label{sec:notations}
\begin{table}[!h]
    \centering
    \caption{Important notations and corresponding descriptions.}
    \begin{tabular}{l|p{6cm}}
        \hline
        \textbf{Notations} & \textbf{Descriptions}  \\
        \hline
        $T$ & Total number of learning tasks\\
        $t$ & Indices of tasks\\
        $\tau$ & Length of time intervals in general\\
        $\mathcal{D}_t^S, \mathcal{D}_t^V, \mathcal{D}_t^Q$ & Support/Validation/Query set of data $\mathcal{D}_t$\\
        $\boldsymbol{\theta}_t, \boldsymbol{\lambda}_t$ & Meta-level primal/dual parameters at round $t$\\
        $\boldsymbol{\theta}_{t,I}, \boldsymbol{\lambda}_{t,I}$ & Interval-level primal/dual parameters for an expert $E_I$ at round $t$\\
        $f_t(\cdot)$ & Loss function at round $t$  \\
        $g_i(\cdot)$ & Fairness function \\
        $m$ & Total number of fairness notions\\
        $i$ & Indices of fairness notions\\
        $\mathcal{G}(\cdot)$ & Base learner\\
        $\mathcal{U}$ & Expert set\\
        $\mathcal{A}_t,\mathcal{S}_t$ & Active/Sleeping expert set at round $t$\\
        $\mathcal{I}$ & AGC interval set\\
        $\mathcal{C}_t$ & Target set of intervals at round $t$\\
        $\mathcal{B}$ & Relaxed primal domain \\
        $\prod_{\mathcal{B}}$ & Projection operation onto domain $\mathcal{B}$\\
        $\eta_1, \eta_2$ & Learning rates\\
        $p_{t,I}$ & Expert weight of $E_I$ at round $t$\\
        $\delta$ & Augmented constant\\
        \hline
    \end{tabular}
    \label{tab:notation}
\end{table}

\subsection{Data Pre-Processing}
\label{app:Data Pre-Processing}
The \textit{New York Stop-and-Frisk} dataset \cite{Koh-icml-2021} is provided by the New York Police Department and contains information about the stop, question, and frisk policy implemented by the NYPD. The full dataset contains records of this policy dating from 2003 to 2019. It includes a record for each stop made, with information about the time of the stop, the location of the stop, information about the officer, the suspect, and other various features. The data used in this paper is from the year 2011, which saw the highest number of stops of any year. This data is used to attempt to predict whether someone will get frisked based on the circumstances of the stop and the features of the individual suspect. Specifically, we extract relevant features (predictor and response) that are utilized for analysis and remove those that may result in data leakage. Missing entries are imputed using precint (\textit{pct}) data and further we remove the \textit{pct} column due to difficulty in binning them into separate pricints. Data records containing unknown or inconsistent values are dropped.

The \textit{MovieLens} dataset was collected through the MovieLens web site (movielens.umn.edu) during the seven-month period from September 19th, 1997 through April 22nd, 1998. This data has been cleaned up - users who had less than 20 ratings or did not have complete demographic information were removed from this data set. We first merge the main three spreadsheets by its primary keys: \textit{user-id} and \textit{item-id}. Then some unused features are removed including \textit{user-id, movie-title, release-date, video-release-date, IMDb-URL} and \textit{zip-code}.

\subsection{Implementation Details and Hyperparameter Tuning}
\label{app:Implementation Details and Hyperparameter Tuning}
\begin{sloppypar}
Our neural network trained follows the same architecture used in \cite{Finn-ICML-2017-(MAML)}, which contains 2 hidden layers of size of 40 with ReLU activation functions. In the training process of the \textit{MovieLens (New York Stop-and-Frisk)} data, each gradient is computed using a batch size of 200 (800) examples where each binary class contains 100 (400) examples. For each dataset, we tune the folowing hyperparameters: (1) the initial dual meta parameter $\boldsymbol{\lambda}_0$ is chosen from $\{$0.00001, 0.0001, 0.001, 0.01, 0.1, 1, 10, 100, 1000, 10000 $\}$; (2) the interval-level gradient steps are chosen from 1 to 10; (3) the number of iterations $N_{meta}$ are chosen from $\{$20, 25, 30, 35, 40, 45, 50, 55, 60, 65, 70, 75, 80, 85, 90, 95, 100$\}$; (4) learning rates $\eta_1$ and $\eta_2$ for updating meta-level parameters in Eq.(\ref{eq:outer-pd-update}) and (15) are chosen from $\{$0.0001, 0.0005, 0.001, 0.005, 0.01, 0.05, 0.1, 0.5, 1, 5, 10, 50, 100, 500, 1000$\}$; (5) the positive constant $\delta$ used in the augmented term are chosen from $\{$10, 25, 50, 75, 100$\}$. 
\end{sloppypar}

\subsection{Additional Results}
\label{sec:additional result}

\begin{figure*}
        \centering
        \begin{subfigure}[b]{0.245\textwidth}
            \includegraphics[width=\textwidth]{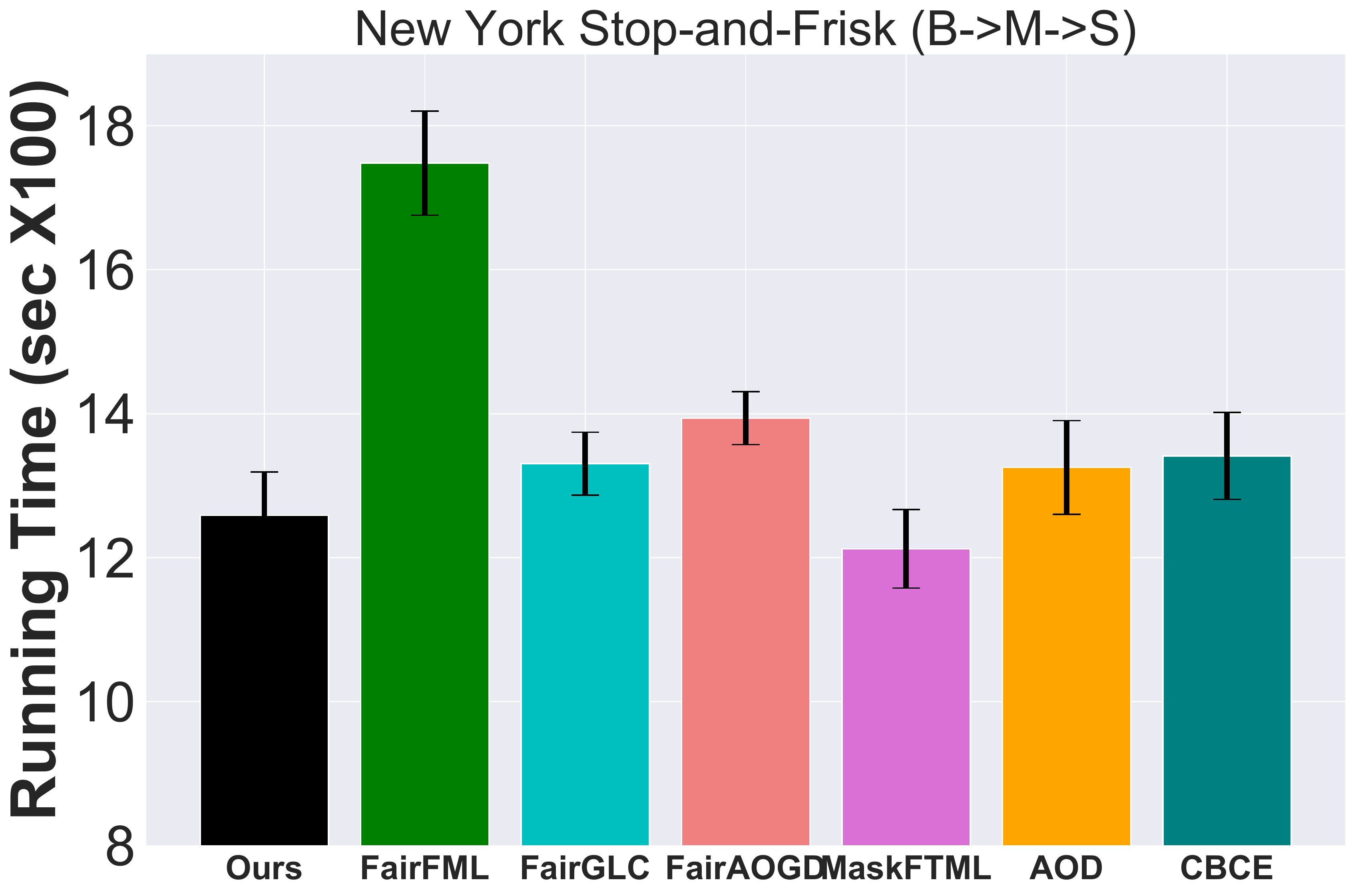}
        \end{subfigure}
        \begin{subfigure}[b]{0.245\textwidth}
            \includegraphics[width=\textwidth]{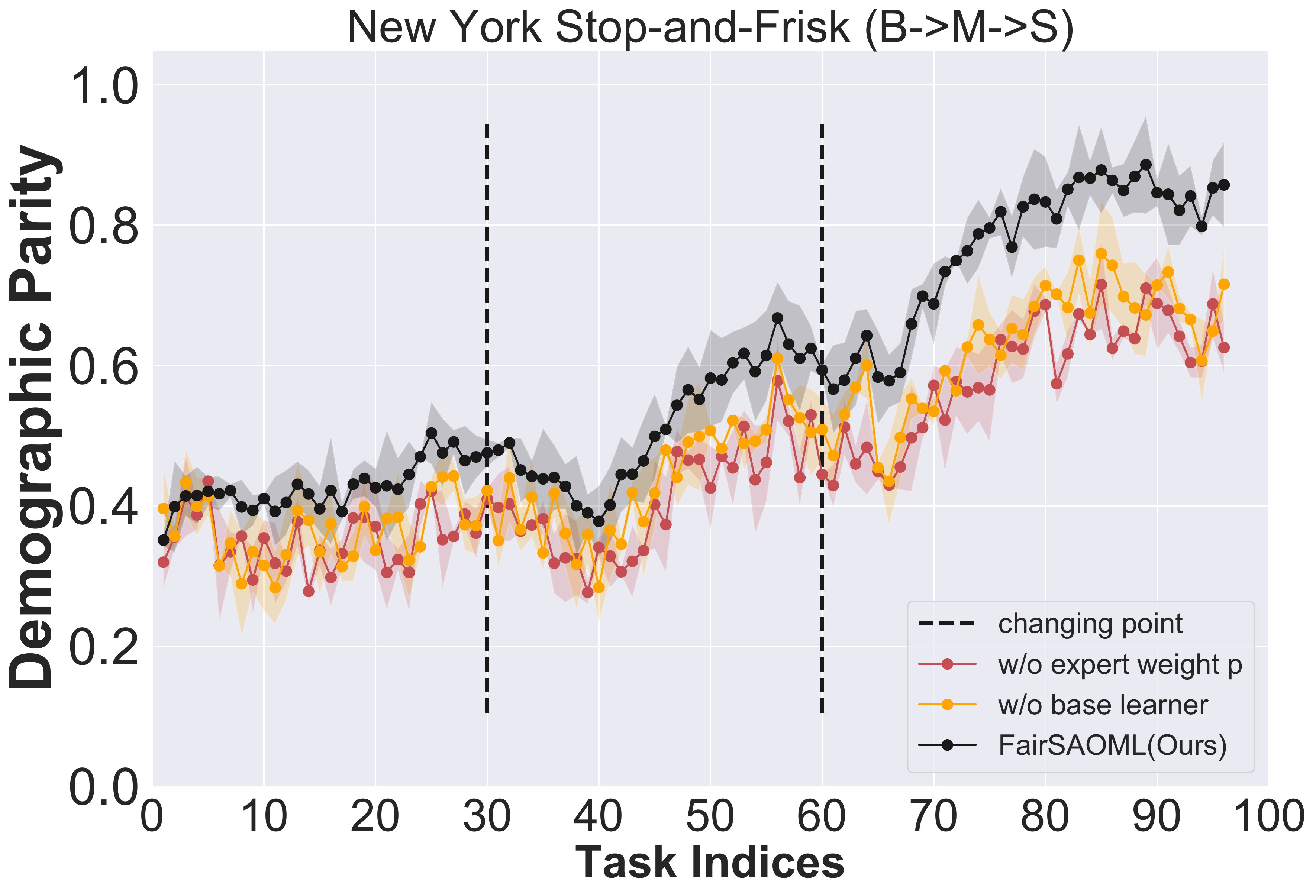}
        \end{subfigure}
        \begin{subfigure}[b]{0.245\textwidth}
            \includegraphics[width=\textwidth]{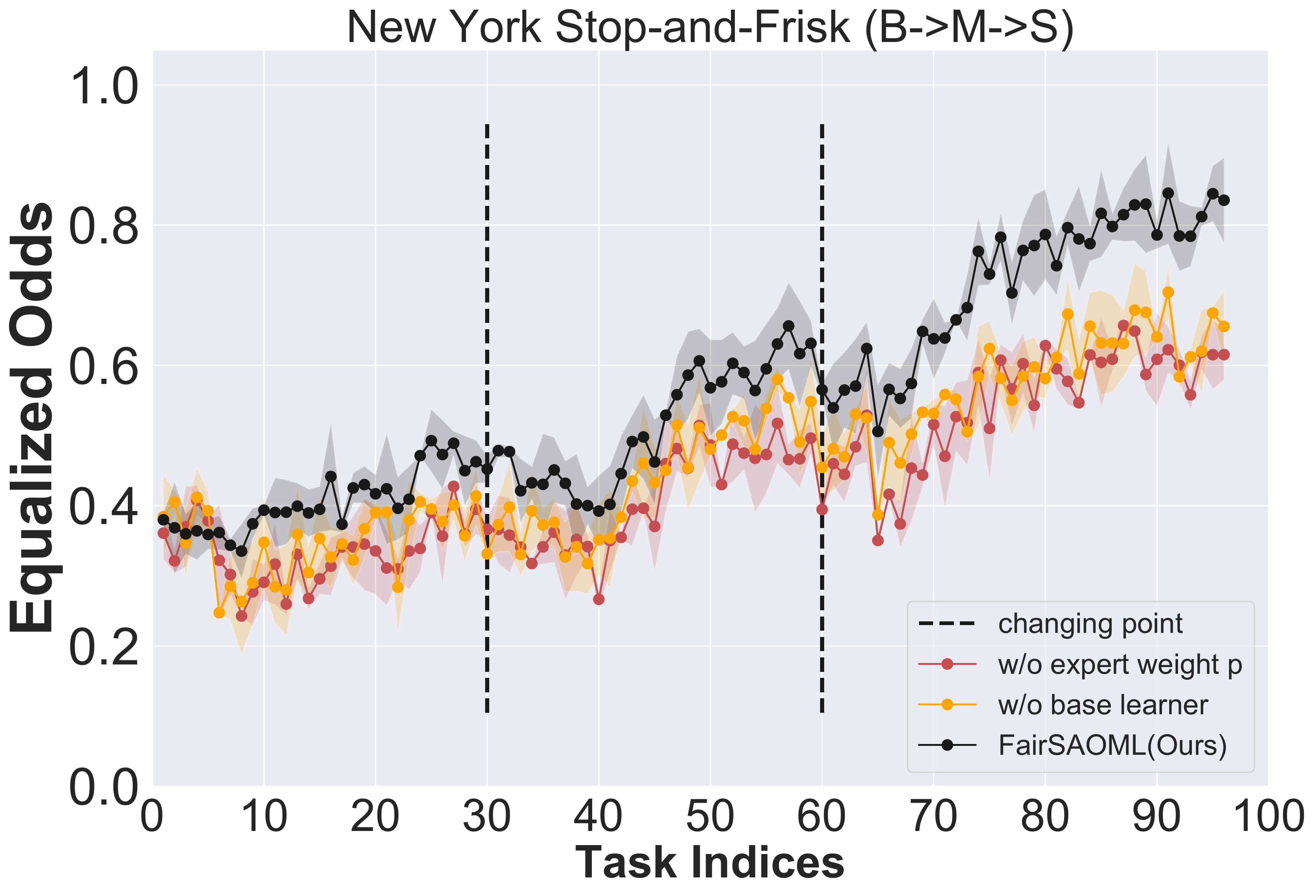}
        \end{subfigure}
        \begin{subfigure}[b]{0.245\textwidth}
            \includegraphics[width=\textwidth]{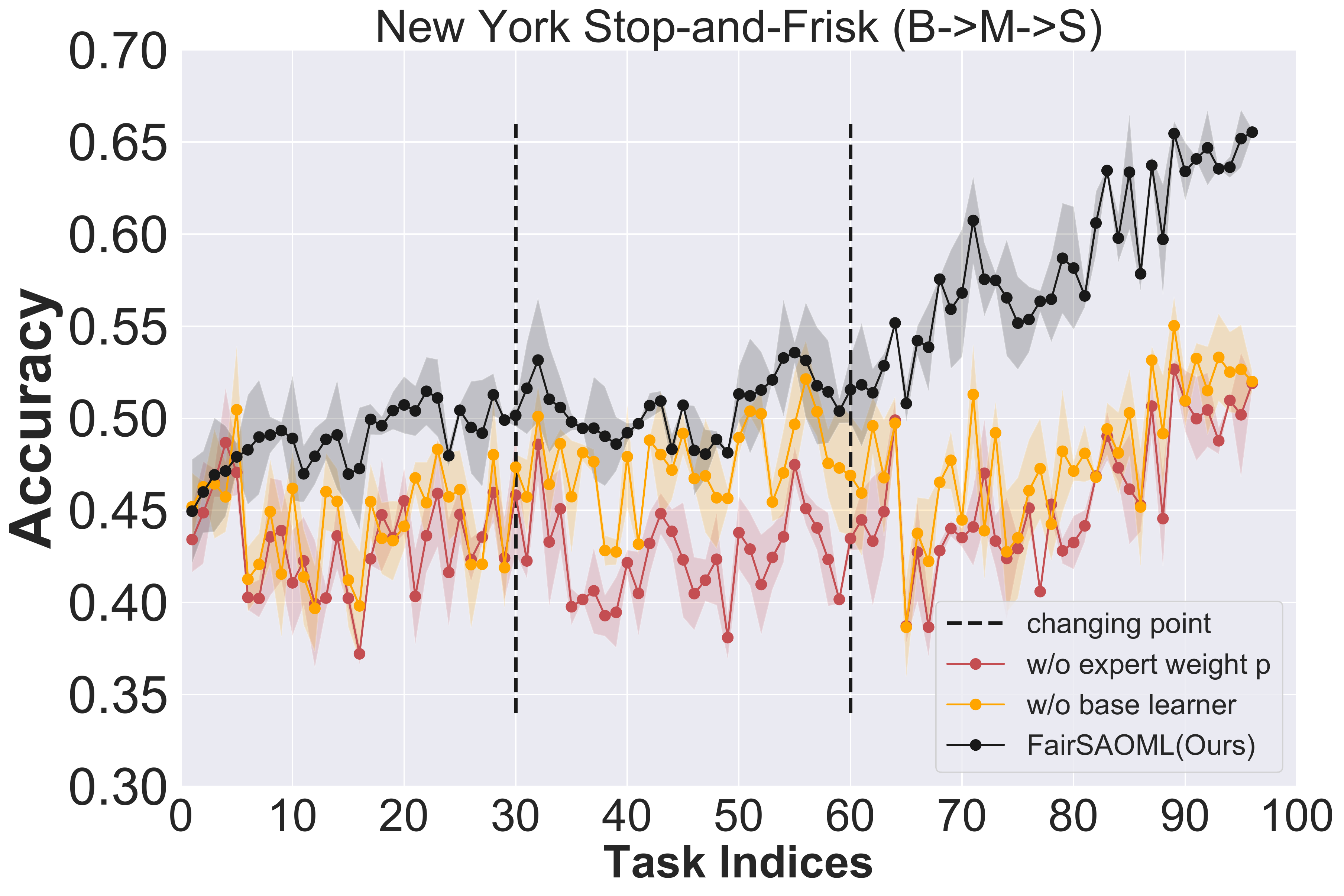}
        \end{subfigure}
        \vspace{-3mm}
        \caption{Running time comparison and ablation studies on the New York Stop-and-Frisk (\textbf{B}$\rightarrow$\textbf{M}$\rightarrow$\textbf{S}) dataset.}
        \label{fig:bms-ablation}
        \vspace{-3mm}
\end{figure*}

\begin{figure*}
        \centering
        \begin{subfigure}[b]{0.245\textwidth}
            \includegraphics[width=\textwidth]{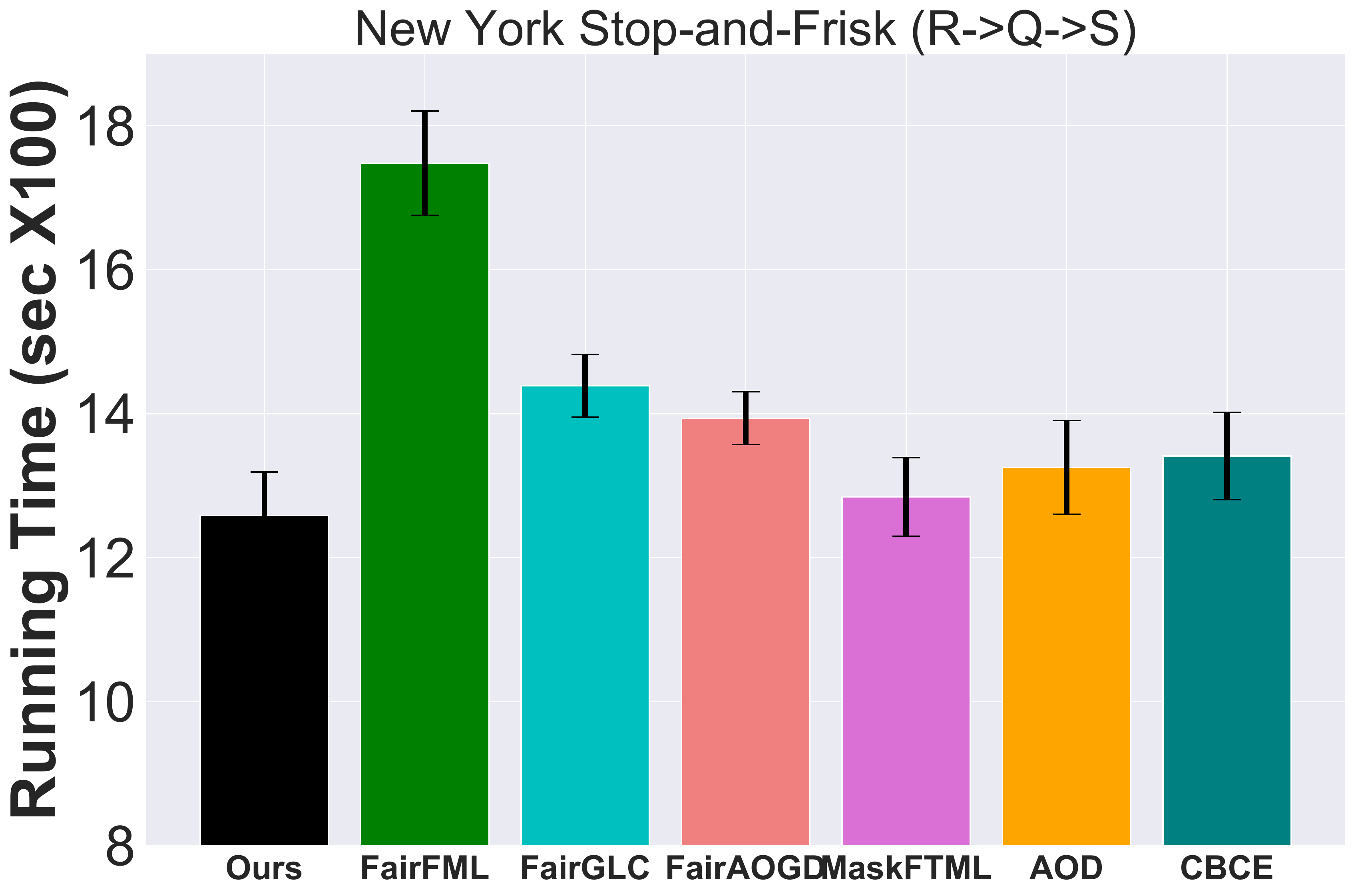}
        \end{subfigure}
        \begin{subfigure}[b]{0.245\textwidth}
            \includegraphics[width=\textwidth]{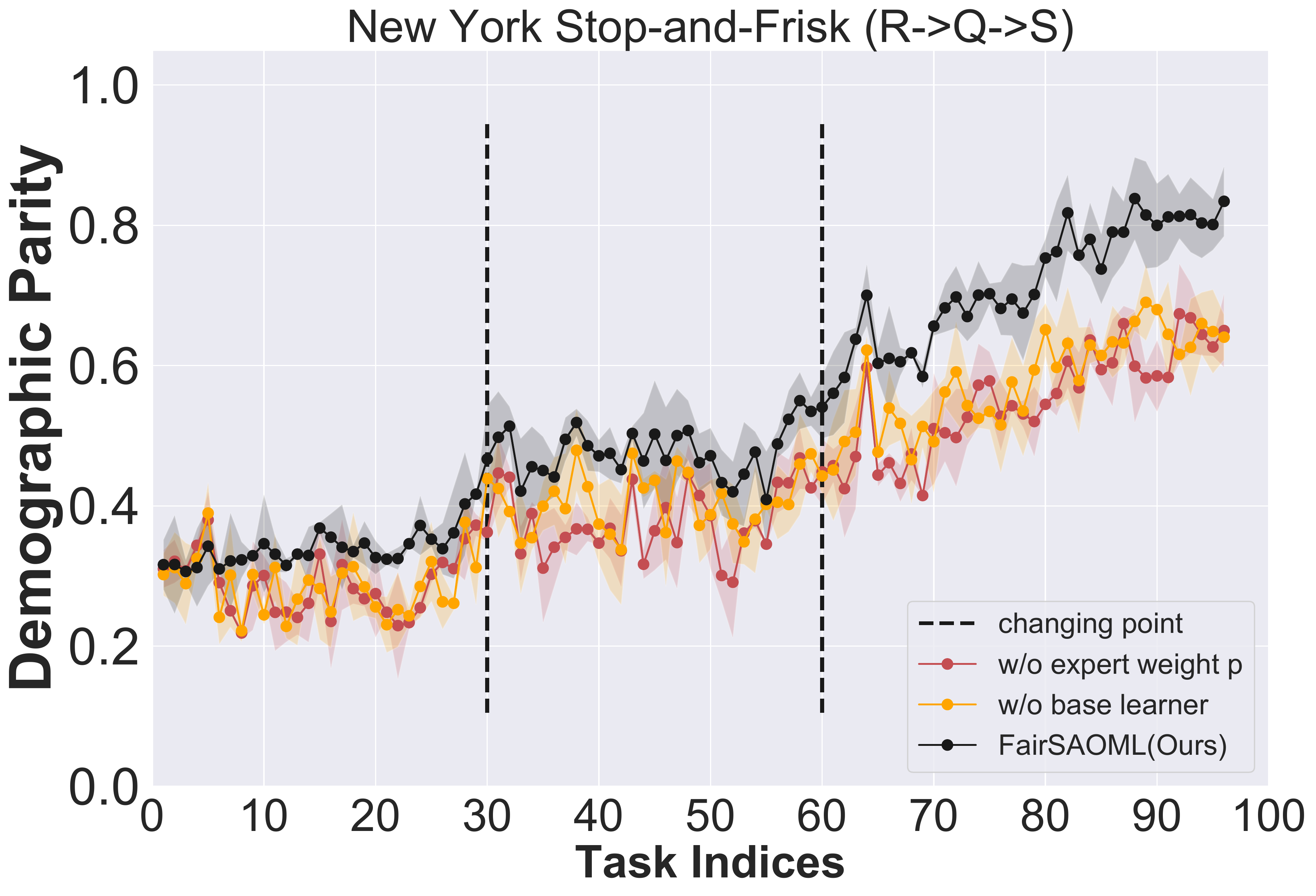}
        \end{subfigure}
        \begin{subfigure}[b]{0.245\textwidth}
            \includegraphics[width=\textwidth]{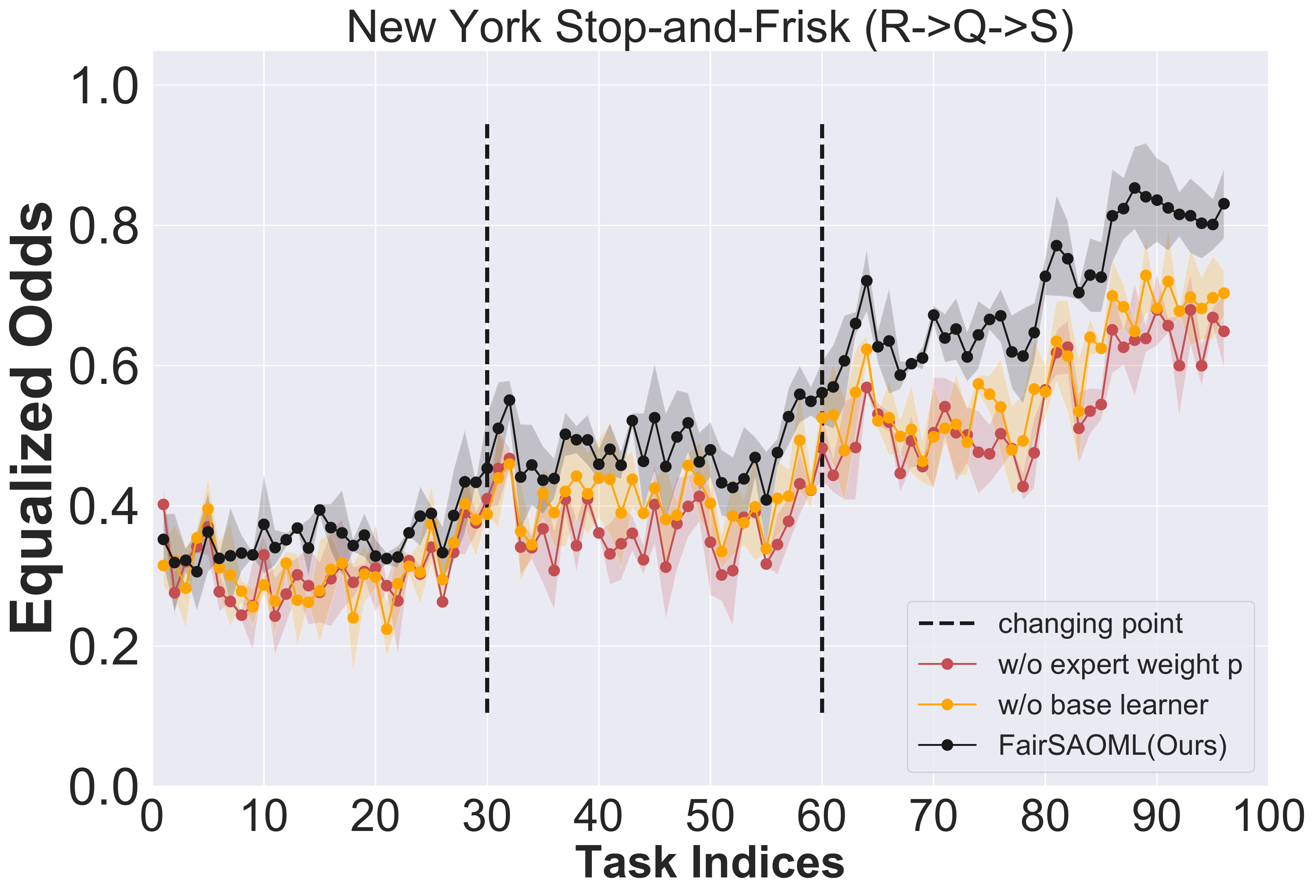}
        \end{subfigure}
        \begin{subfigure}[b]{0.245\textwidth}
            \includegraphics[width=\textwidth]{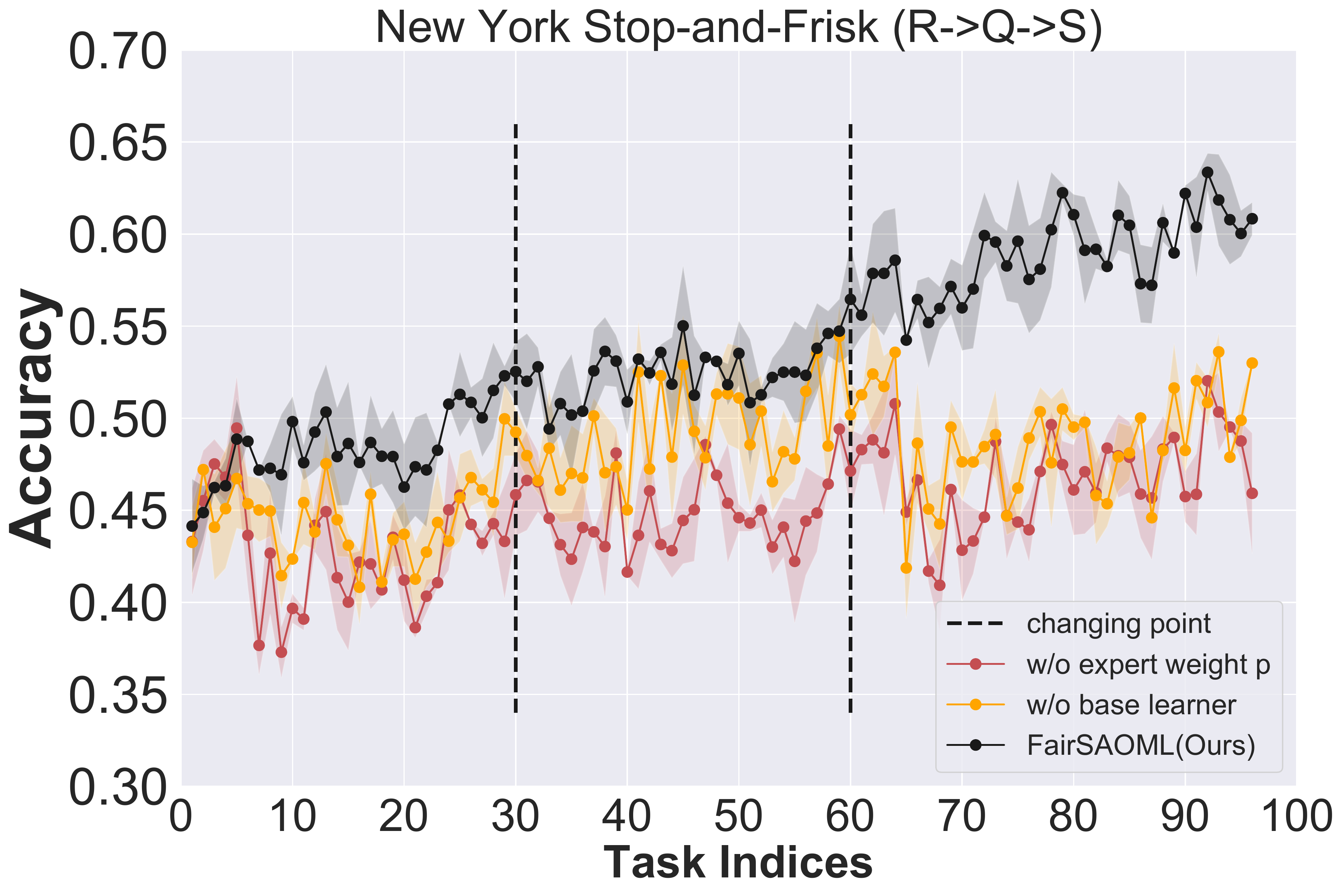}
        \end{subfigure}
        \vspace{-3mm}
        \caption{Running time comparison and ablation studies on the New York Stop-and-Frisk (\textbf{R}$\rightarrow$\textbf{Q}$\rightarrow$\textbf{S}) dataset.}
        \label{fig:rqs-ablation}
        \vspace{-3mm}
\end{figure*}

Due to space limit, more experimental results on the \textit{New York Stop-and-Frisk} dataset are given in Figure \ref{fig:bms-ablation} and \ref{fig:rqs-ablation}. 

\section{Proof sketch of Theorem \ref{main theorem}}
\label{App:proof}
    Under Assumptions \ref{assmp1}, \ref{assmp2} and \ref{assmp3}, we first target Eq.(\ref{eq:L_t}) and have

\begin{lemma}[Theorem 1 in \cite{zhao-KDD-2021}]
\label{lemma:convex-concave}
\begin{sloppypar}
    Suppose $f$ and $g:\Theta\times\mathbb{R}_+^m \rightarrow \mathbb{R}$ satisfy Assumptions \ref{assmp1}, \ref{assmp2} and \ref{assmp3}. The interval-level update and the augmented Lagrangian function $\mathcal{L}_t(\boldsymbol{\theta, \lambda})$ are defined in Eq.(\ref{eq:inner-pd-update})(12) and Eq.(\ref{eq:L_t}). Then, the function $\mathcal{L}_t(\boldsymbol{\theta, \lambda})$ is convex-concave with respect to the arguments $\boldsymbol{\theta}$ and $\boldsymbol{\lambda}$, respectively. Furthermore, as for $\mathcal{L}_t(\boldsymbol{\cdot,\lambda})$, if stepsize $\eta_{t,I}$ for each active expert $E_I$ is selected as $\eta_{t,I}\leq\min\{\frac{\mu_f+\Bar{\lambda}m\mu_g}{8(L_f+\Bar{\lambda}mL_g)(\rho_f+\Bar{\lambda}m\rho_g)}, \frac{1}{2(\beta_f+\Bar{\lambda}m\beta_g)}\}$, then $\mathcal{L}_t(\boldsymbol{\cdot,\lambda})$ enjoys $\frac{9}{8}(\beta_f+\Bar{\lambda}m\beta_g)$-smooth and $\frac{1}{8}(\mu_f+\Bar{\lambda} m\mu_g)$-strongly convex, where $\Bar{\lambda}\geq 0$ is the mean value of $\boldsymbol{\lambda}$.
\end{sloppypar}
\end{lemma}

According to Theorems 1 and 3 in \cite{luo-2015-achieving} and the Lemma 1 in \cite{zhang-2020-AISTATS}, we have the following lemma with respect to Eq.(\ref{eq:L_t}) that
\begin{lemma}
\label{lemma:meta-regret}
Under Assumption \ref{assmp2}, for any interval $I=[i,j]\in\mathcal{I}$, \sysname{} satisfies
\begin{align*}
    \sum_{u=i}^t \mathcal{L}_u(\boldsymbol{\theta}_u,\boldsymbol{\lambda}_u)-\sum_{u=i}^t \mathcal{L}_u(\boldsymbol{\theta}_{u,I},\boldsymbol{\lambda}_{u,I}) \leq S\sqrt{6L_f L_g(t-i-1)c(t)}
\end{align*}
where $c(t)\leq 1+\ln t + \ln(1+\log_2^T)+\ln\frac{5+3\ln (1+t)}{2}$.
\end{lemma}

By applying the Lemma \ref{lemma:meta-regret} with the Theorem 2 in \cite{zhang-2020-AISTATS}, we have
\begin{lemma}
\label{lemma:regret-single-interval}
    Under Assumption \ref{assmp1} and \ref{assmp2}, for any interval $I=[i,j]\in\mathcal{I}$, for any $(\boldsymbol{\theta,\lambda})\in\Theta\times\mathbb{R}^m_+$ \sysname{} satisfies
    \begin{align*}
        \sum_{t\in I} \mathcal{L}_t(\mathcal{G}_t(\boldsymbol{\theta}_t),\boldsymbol{\lambda})-
        \sum_{t\in I} \mathcal{L}_t(\boldsymbol{\theta},\boldsymbol{\lambda}_{t,I})
        \leq S\sqrt{|I|}(\sqrt{6L_f L_g c(t)}+G)
    \end{align*}
\end{lemma}

To extend our Lemma \ref{lemma:regret-single-interval} to any interval $I=[r,s]\subseteq [T]$, we refer the following lemma
\begin{lemma}[Lemma 3 in \cite{zhang-2020-AISTATS}]
\label{lemma:interval patitions}
    For any interval $[r,s]\subseteq[T]$, it can be partitioned into two sequences of disjoint and consecutive intervals, denoted by $I_{-p},...,I_0\in\mathcal{I}$ and $I_1,...,I_q\in\mathcal{I}$, such that
    \begin{align*}
        |I_{-i}|/|I_{-i+1}|\leq 1/2, \forall i\geq 1 \quad \textit{and} \quad |I_i|/|I_{i-1}|\leq 1/2, \forall i\geq 2
    \end{align*}
\end{lemma}

Next, we prove Theorem \ref{main theorem}.
\begin{proof}
By applying Lemma \ref{lemma:regret-single-interval} onto Lemma \ref{lemma:interval patitions} and set $\boldsymbol{\theta}^*$ being the optimal solution for $\min_{\boldsymbol{\theta}\in\Theta} \sum_{t=r}^s f_t(\mathcal{G}_t(\boldsymbol{\theta}))$ where $[r,s]\subseteq[T]$, we have
\begin{align}
\label{eq:L_t regret}
    &\sum_{t=r}^s \mathcal{L}_t(\mathcal{G}_t(\boldsymbol{\theta}_t),\boldsymbol{\lambda})- \sum_{t=r}^s \mathcal{L}_t(\mathcal{G}_t(\boldsymbol{\theta}^*),\boldsymbol{\lambda}_{t,I}) \nonumber\\
    =&\sum_{i=-p}^q \Big( \sum_{t\in I_i} \mathcal{L}_t(\mathcal{G}_t(\boldsymbol{\theta}_t),\boldsymbol{\lambda})- \sum_{t\in I_i} \mathcal{L}_t(\mathcal{G}_t(\boldsymbol{\theta}^*),\boldsymbol{\lambda}_{t,I})\Big) \nonumber\\
    \leq&\sum_{i=-p}^q S\sqrt{|I_i|}(\sqrt{6L_f L_g c(s)}+G)\nonumber\\
    \leq& 2S(\sqrt{6L_f L_g c(s)}+G)\sum_{i=0}^{\infty} \sqrt{2^{-i}|I|}\nonumber\\
    \leq& 8S(\sqrt{6L_f L_g c(s)}+G) \sqrt{|I|}
\end{align}

By expanding Eq.(\ref{eq:L_t regret}) using Eq.(\ref{eq:L_t}) and following the Theorem 3.1 in \cite{Cambridge-book-2006}, we have
\begin{align*}
    &\sum_{t=r}^s\Big\{f_t(\mathcal{G}_t(\boldsymbol{\theta}_t))-f_t(\mathcal{G}_t(\boldsymbol{\theta}^*))\Big\} \\
    &+\sum_{i=1}^m\Big\{\lambda_i\sum_{t=r}^s g_i(\mathcal{G}_t(\boldsymbol{\theta}_t))-\sum_{t=r}^s\lambda_{t,i}g_i(\mathcal{G}_t(\boldsymbol{\theta}^*))\Big\}\\
    &-\frac{\delta(\eta_1+\eta_2) (s-r+1)}{2}||\boldsymbol{\lambda}||^2+\frac{\delta(\eta_1+\eta_2)}{2}\sum_{t=r}^s||\boldsymbol{\lambda}||^2 \\
    &\leq 8S \Big(\sqrt{6L_f L_g c(s)}+G\Big) \sqrt{|I|}
\end{align*}
\begin{sloppypar}
Here, we approximately average $p_{t,I}$ for all experts $E_I$ in $\mathcal{U}$ and hence the subscription $I$ is omitted. 
Inspired by the proof of Theorem 4 in \cite{OGDLC-2012-JMLR}, we take maximization for $\boldsymbol{\lambda}$ over $(0,+\infty)$ and get
\end{sloppypar}
\begin{align*}
    &\sum_{t=r}^s\Big\{f_t(\mathcal{G}_t(\boldsymbol{\theta}_t))-f_t(\mathcal{G}_t(\boldsymbol{\theta}^*))\Big\} \\
    &+\sum_{i=1}^m\Big\{\frac{\big[\sum_{t=r}^s g_i(\mathcal{G}_t(\boldsymbol{\theta}_t))\big]^2_+}{2(\delta(\eta_1+\eta_2)(s-r+1)+\frac{m}{\eta_1+\eta_2})}-\sum_{t=r}^s\lambda_{t,i}g_i(\mathcal{G}_t(\boldsymbol{\theta}^*))\Big\}\\
    &\leq 8S\Big(\sqrt{6L_f L_g c(s)}+G\Big) \sqrt{|I|}
\end{align*}
Since $g_i(\mathcal{G}_t(\boldsymbol{\theta}^*))\leq 0$ and $\lambda_{t,i}\geq 0, \forall i\in[m]$, the resulting inequality becomes
\begin{align*}
    &\sum_{t=r}^s\Big\{f_t(\mathcal{G}_t(\boldsymbol{\theta}_t))-f_t(\mathcal{G}_t(\boldsymbol{\theta}^*))\Big\} \\
    &+\sum_{i=1}^m\frac{\big[\sum_{t=r}^s g_i(\mathcal{G}_t(\boldsymbol{\theta}_t))\big]^2_+}{2(\delta(\eta_1+\eta_2)(s-r+1)+\frac{m}{\eta_1+\eta_2})}\leq 8S\Big(\sqrt{6L_f L_g c(s)}+G\Big) \sqrt{|I|}
\end{align*}
Due to non-negative of $\frac{\big[\sum_{t=r}^s g_i(\mathcal{G}_t(\boldsymbol{\theta}_t))\big]^2_+}{2(\delta(\eta_1+\eta_2)(s-r+1)+\frac{m}{\eta_1+\eta_2})}$, we have
\begin{align*}
    \sum_{t=r}^s\Big\{f_t(\mathcal{G}_t(\boldsymbol{\theta}_t))-f_t(\mathcal{G}_t(\boldsymbol{\theta}^*))\Big\} &\leq 8S\Big(\sqrt{6L_f L_g c(s)}+G\Big) \sqrt{|I|}\\
    &=O\Big((|I|\log s)^{1/2}\Big)
\end{align*}
Furthermore, we have $\sum_{t=r}^s\Big\{f_t(\mathcal{G}_t(\boldsymbol{\theta}_t))-f_t(\mathcal{G}_t(\boldsymbol{\theta}^*))\Big\}\geq-F(s-r+1)$ according to the assumption and set $\eta_1=\eta_2=O(1/\sqrt{s})$.
We have
\begin{align*}
    \sum_{t=r}^s g_i(\mathcal{G}_t(\boldsymbol{\theta}_t))\leq O\Big((|I|s\log s)^{1/4}\Big)
\end{align*}
Therefore, as for FairSAR proposed in Eq.(\ref{eq:ourRegret}) we complete the proof
\begin{align*}
    &\max_{[s,s+\tau-1]\subseteq[T]} \bigg( \sum_{t=s}^{s+\tau-1} f_t\Big(\mathcal{G}_t(\boldsymbol{\theta}_t)\Big) - f_t\Big(\mathcal{G}_t(\boldsymbol{\theta}^*)\Big) \bigg) \leq O\Big((\tau\log T)^{1/2}\Big)\\
    &\max_{[s,s+\tau-1]\subseteq[T]} \bigg (\sum_{t=s}^{s+\tau-1} g_i\Big(\mathcal{G}_t(\boldsymbol{\theta}_t)\Big) \bigg)\leq O\Big((\tau T\log T)^{1/4}\Big), \quad \forall i \in[m]
\end{align*}

\end{proof}

\end{document}